\documentclass{article}

\PassOptionsToPackage{numbers, compress}{natbib}
\usepackage[preprint]{neurips_2026}

\usepackage[utf8]{inputenc} 
\usepackage[T1]{fontenc}    
\usepackage{hyperref}       
\usepackage{url}            
\usepackage{booktabs}       
\usepackage{multirow}       
\usepackage{amsmath}        
\usepackage{amsfonts}       
\usepackage{nicefrac}       
\usepackage{microtype}      
\usepackage{xcolor}         
\usepackage{graphicx}       
\usepackage{wrapfig}        
\usepackage{float}          
\usepackage{algorithm}      
\usepackage{algpseudocode}  
\usepackage[most]{tcolorbox} 
\usepackage{enumitem}        

\newtcolorbox[auto counter]{observation}[1][]{%
  enhanced, breakable,
  colback={rgb,255:red,240;green,240;blue,240}, 
  colframe={rgb,255:red,160;green,160;blue,160}, 
  fontupper=\small,
  boxrule=0.5pt, arc=2pt,
  left=8pt, right=8pt, top=4pt, bottom=4pt,
  before upper={\textbf{Observation~\thetcbcounter.}\quad},
  #1
}

\newtcolorbox[auto counter]{takeaway}[1][]{%
  enhanced, breakable,
  colback={rgb,255:red,240;green,240;blue,240}, 
  colframe={rgb,255:red,160;green,160;blue,160}, 
  fontupper=\small,
  boxrule=0.5pt, arc=2pt,
  left=8pt, right=8pt, top=4pt, bottom=4pt,
  before upper={\textbf{Takeaway~\thetcbcounter.}\quad},
  #1
}

\newcommand{\CGR}{CGR}
\newcommand{\VAR}{VAR}
\newcommand{\AGLA}{AGLA}

\title{Magnifying What Matters: Attention-Guided Adaptive Rendering for Visual Text Comprehension}

\author{%
  Shenglai Zeng$^{1}$\quad
  Qirui Wang$^{2}$\quad
  Kai Guo$^{1}$\quad
  Xinnan Dai$^{1}$\quad
  Xianxuan Long$^{1}$\quad
  Hui Liu$^{1}$ \\[4pt]
  $^{1}$Michigan State University\qquad
  $^{2}$Xi'an Jiaotong University \\
}

\begin{document}

\maketitle

\begin{abstract}
  Visual Text Comprehension (VTC) renders text into images for a
vision--language model (VLM) to read, sidestepping LLM context-window
limits and powering applications from long-page OCR to multi-page memory
QA. Yet
existing VTC pipelines treat rendering and layout as a fixed,
content-agnostic preprocessing step, and offer little mechanistic
understanding of how VLMs internally process visualized text. Through a
focused empirical study on VTC QA tasks, we reveal that VLMs exhibit a
\emph{localization-without-utilization} regime: evidence-localizing
attention emerges sharply in the middle-to-late layers and is largely
decoupled from answer correctness, yet simply enlarging the localized
spans on the rendered page recovers a large fraction of the failures.
Building on these observations, we propose \textbf{AGAR}
(\underline{A}ttention-\underline{G}uided \underline{A}daptive
\underline{R}endering), a training-free, model-agnostic method that
leverages a VLM's own middle-to-late layer attention to identify the
top-$K$ important visual patches, maps them back to word spans, and
re-renders the page with those spans enlarged before re-inferring the
answer. Extensive experiments across nine VTC benchmarks (short-form,
long-context, and multi-page memory QA) and four VLM backbones show that
AGAR (i)~consistently improves off-the-shelf VLMs as a plug-and-play
enhancement, (ii)~composes with VLM post-training to yield further gains,
and (iii)~remains robust under both visual- and text-side input
degradation.

\end{abstract}

\section{Introduction}
\label{sec:intro}

The context window of large language models (LLMs) remains a fundamental
bottleneck that limits how much text a model can process in a single
pass. A recently emerging paradigm, \emph{Visual Text
Comprehension} (VTC) or \emph{optical context compression}, sidesteps this
bottleneck by \emph{rendering} text into images and letting a
vision--language model (VLM) read it directly~\citep{wei2025deepseek,
wei2026deepseek, cheng2025glyph}. A single visual patch can pack many
characters, yielding $3\!-\!20\times$ token-level compression and
unlocking applications such as long-page
OCR~\citep{wei2025deepseek}, long-context visual question
answering~\citep{cheng2025glyph}, vision-based long-term memory for
agents~\citep{shi2026memocr, feng2026agentocr}, reasoning-chain
compression~\citep{wang2026vtc}, persona-conditioned
recommendation~\citep{qiao2026text}, and long-context code
understanding~\citep{zhong2026can}.

Despite this momentum, current VTC pipelines leave two important gaps that remain underexplored.
First, most prior work treats rendering and layout as a fixed, content-agnostic step. But typography itself is a useful signal. Textbooks enlarge headings and bold keywords to direct a human reader's eye~\citep{rayner1998eye,hyona2004effects}, and similar visual emphasis could possibly route a VLM's limited visual capacity toward task-relevant content.
Second, the community has limited \emph{mechanistic} understanding of how VLMs internally process VTC inputs~\citep{olsson2022context,ben2024lvlm}: which layers actually localize task-relevant text, whether failures stem from attending to the wrong region or from mis-reading the right one, and how the model consumes the localized signal. Understanding these internal mechanisms is crucial for diagnosing bottlenecks and designing effective enhancement methods. By contrast, in adjacent areas such as evidence-based question answering and traditional VQA, prior work has shown that the internal representations and attention patterns of LLMs already contain rich, exploitable signals about what the model is truly attending to~\citep{zeng2025towards,zeng2026attn,liu2025selfelicit,liu2025seeing}, suggesting that a similar internal lens may be the missing tool for VTC as well.


To investigate the two gaps above, we conduct a comprehensive empirical
study on VTC QA tasks (\S\ref{sec:preliminary}) that examines how a VLM
reads long rendered text and where its failures come from, looking at
both its attention patterns and its answer quality. We make three key observations. \textbf{First}, evidence-localizing
attention emerges sharply in the middle-to-late layers and is carried
by a substantial fraction of heads in those layers, consistently across
backbones.
\textbf{Second}, this localization is largely decoupled from correctness:
incorrect samples attend to the right evidence nearly as well as correct
ones, placing VLMs in a \emph{localization-without-utilization} regime.
\textbf{Third}, on failing samples, enlarging the ground-truth evidence
on the rendered page recovers a large fraction of errors with a high
fix-to-break ratio. These findings suggest that, at least to a large extent, the
bottleneck lies in \emph{utilizing} the localized evidence rather than
discovering it.

Inspired by the above observations, we develop \textbf{AGAR}
(\underline{A}ttention-\underline{G}uided \underline{A}daptive
\underline{R}endering), a training-free, model-agnostic wrapper around any
VLM that exposes layer-wise attentions. AGAR first leverages middle-to-late
layer attention scores to identify the top-$K$ important visual patches
and maps them back to word spans; it then re-renders the page with those
spans enlarged and re-infers the answer. Extensive experiments across
nine VTC benchmarks spanning short-form, long-context, and multi-page
memory QA, and four VLM backbones (Qwen3-VL-8B~\citep{qwen3technicalreport,Qwen2.5-VL},
InternVL3.5-8B~\citep{wang2025internvl3},
GLM-4.1V-9B-Thinking~\citep{hong2025glm}, and
Glyph~\citep{cheng2025glyph}), show that AGAR is
\textbf{(i)}~training-free, raising VTC accuracy on off-the-shelf VLMs as
a plug-and-play enhancement; \textbf{(ii)}~complementary to post-training,
yielding additional gains when its attention-guided magnification is
incorporated into supervised fine-tuning; and \textbf{(iii)}~robust to
both visual-side (lower resolution, font/style perturbations) and
text-side (distractor injection, noisy contexts) input degradation.

In summary, our contributions are:
\begin{itemize}[leftmargin=*, nosep]
    \item We identify a \emph{localization-without-utilization} regime in
    VTC: VLMs already attend to the right evidence in the middle-to-late
    layers even when the answer is incorrect, yet enlarging those spans
    on the rendered page recovers many failures, pinpointing evidence
    \emph{utilization} as a key bottleneck.
    \item We propose \textbf{AGAR}, an attention-guided adaptive rendering
    method that closes this gap by using the model's own attention to
    select evidence spans and enlarging them in a re-rendered page, with
    no weight or prompt changes.
    \item Extensive experiments across 9 benchmarks and 12 subtasks
    on four VLM backbones show
    that AGAR (a) improves off-the-shelf VLMs in a training-free manner,
    (b) composes with VLM post-training to yield further gains, and (c)
    is robust to both visual- and text-side input degradation.
\end{itemize}

\section{Related Works}
\label{sec:related}

\subsection{Visual Text Comprehension}
\label{sec:related:vtc}

Visual Text Comprehension (VTC), or optical context compression, sidesteps
the LLM context-window bottleneck by rendering text into images and
letting a VLM read it directly~\citep{wei2025deepseek, wei2026deepseek,
cheng2025glyph, zhao2025vtcbench, liu2026vista}. Building on this
paradigm, a growing body of work targets
downstream applications including vision-based long-term agent
memory~\citep{shi2026memocr, feng2026agentocr}, long reasoning-chain
compression~\citep{wang2026vtc}, long-context code
understanding~\citep{zhong2026can}, and generative
recommendation~\citep{qiao2026text}. Most of these works, however, either
treat the rendering pipeline as a fixed, content-agnostic preprocessing
step, or train a dedicated model to pre-compress the context end-to-end
\citep{shi2026memocr}; a clear gap remains in understanding how VLMs
internally process VTC inputs and in exploiting such mechanistic signals
to drive context-aware rendering.

\subsection{Utilization of (V)LLMs' Internal Signals}
\label{sec:related:internal}

An emerging research direction focuses on utilizing (V)LLMs' internal
signals to understand how they process input contexts. On the LLM side,
\citet{zeng2025towards2} use hidden representations to identify
high-level concepts in RAG systems such as context helpfulness;
\citet{liu2025selfelicit} use last-token attention to surface important
evidence spans in text QA; and \citet{zeng2026attn} exploit attention
scores to locate critical personalization signals in long user
histories. On the VLM side, prior work in classical VQA has shown that
attention-related signals can be used to amplify task-relevant visual
regions and suppress irrelevant ones, via Grad-CAM-style
attribution~\citep{selvaraju2017grad, an2025mitigating}, attention-driven
masking~\citep{liu2025unveiling}, and brightness/contrast
modulation~\citep{liu2025seeing}. VTC, however, lies at the intersection
of these two regimes---visual inputs that encode dense, structured text
rather than natural-image scenes or pure-text tokens---and has not been
systematically studied through this internal-signal lens.

\section{Preliminary Studies}
\label{sec:preliminary}

Before describing our method, we conduct a focused empirical study
across four open-weight VLMs (Qwen3-VL-8B, InternVL3.5-8B,
GLM-4.1V-9B-Thinking, and Glyph\footnote{Glyph is post-trained on
GLM-4.1V-9B-Thinking for VTC.}) to understand \emph{how} a VLM reads
long rendered text and where its failures come from. After
introducing notation and metrics (\S\ref{sec:prelim:problem}), we
organize the study around three research questions that
progressively dissect the gap between \emph{seeing} and \emph{using}
evidence: \textbf{RQ1}~(\S\ref{sec:prelim:obs1}) asks \emph{where}
evidence-localizing attention emerges inside a VLM;
\textbf{RQ2}~(\S\ref{sec:prelim:obs2}) asks whether VTC failures are
themselves localization failures; \textbf{RQ3}~(\S\ref{sec:prelim:obs3})
asks how to help the model use the evidence it already localizes.
Each subsection ends with a boxed observation; together they motivate
the design of our method (\S\ref{sec:methods}).

\subsection{Problem Formulation}
\label{sec:prelim:problem}

\paragraph{Visual Text Comprehension (VTC).}
Given a long textual context $T$ rendered into one or more images
$\mathcal{I} = \{I_1, \ldots, I_{|\mathcal{I}|}\}$ and a natural-language question $q$, a
vision--language model (VLM) $\mathcal{M}$ produces an answer
$\hat a = \mathcal{M}(\mathcal{I}, q)$. We assume that each ground-truth answer
$a^\star$ is supported by a set of \emph{evidence spans}
$\mathcal{E} \subseteq T$, i.e., the key evidence in $T$ supporting
$a^\star$ (provided as explicit annotations in the QA datasets we
use).

\paragraph{Image patches and evidence labels.}
The VLM's visual encoder partitions $\mathcal{I}$ into a grid of patches and
emits $N$ visual tokens $\mathcal{P} = \{p_1, \ldots, p_N\}$, one per
patch.\footnote{A "patch" is the image region of one visual token (possibly multiple ViT sub-patches after spatial merging).}
Using the renderer's word--bounding-box map together with $\mathcal{E}$,
each patch is tagged as \emph{evidence} or \emph{non-evidence}:
$y_i = 1$ if patch $p_i$ overlaps any rendered region of $\mathcal{E}$,
otherwise $y_i = 0$. The evidence set is denoted as
$\mathcal{P}_{\mathcal{E}} = \{p_i : y_i = 1\}$.

\paragraph{Patch-level attention scores.}
Attention is extracted from the last query token $q_{\text{last}}$ (the
position that drives next-token prediction) to image patches. Let $L$
denote the number of decoder layers and $H$ the heads per layer. For
each layer~$\ell$ and head~$h$, the per-head attention vector is
\[
\boldsymbol{\alpha}^{(\ell, h)} \;=\; \big[\,A^{(\ell, h)}_{q_{\text{last}},\, p_1},\,\ldots,\, A^{(\ell, h)}_{q_{\text{last}},\, p_N}\,\big] \;\in\; \mathbb{R}^N,
\]
where $A^{(\ell, h)}$ is the (post-softmax) self-attention map at layer $\ell$,
head $h$. Two aggregated attention vectors are used downstream.
(i) The \emph{head-mean attention vector} per layer,
$\bar{\boldsymbol{\alpha}}^{(\ell)} = \tfrac{1}{H}\sum_{h=1}^{H} \boldsymbol{\alpha}^{(\ell, h)}$.
(ii) The \emph{layer-range aggregated attention vector} over a contiguous
interval $\mathcal{L}_{[a,b]} = \{\ell : \lceil a L \rceil \leq \ell \leq \lfloor b L \rfloor\}$
with $0 \leq a < b \leq 1$,
\[
\tilde{\boldsymbol{\alpha}}_{[a,b]}
\;=\;
\frac{1}{|\mathcal{L}_{[a,b]}|}
\sum_{\ell \in \mathcal{L}_{[a,b]}} \bar{\boldsymbol{\alpha}}^{(\ell)}.
\]

\paragraph{Localization quality (NDCG).}
Given an attention score vector $\mathbf{s} \in \mathbb{R}^N$  and the binary evidence labels
$\mathbf{y}$, we measure its evidence-localization quality using Normalized
Discounted Cumulative Gain~\citep{jarvelin2002cumulated}.
Sorting $\mathcal{P}$ by $\mathbf{s}$ (NDCG) in descending order with $\pi(r)$
denoting the patch at rank $r$, we define
$\mathrm{DCG}(\mathbf{s}, \mathbf{y}) = \sum_{r=1}^{N} y_{\pi(r)} / \log_2(r+1)$
and
$\mathrm{NDCG}(\mathbf{s}, \mathbf{y}) = \mathrm{DCG}(\mathbf{s}, \mathbf{y}) / \mathrm{IDCG}(\mathbf{y})$,
where $\mathrm{IDCG}$ is the DCG of the ideal ranking that places all
$|\mathcal{P}_{\mathcal{E}}|$ positives first. NDCG~$=1$ corresponds to a
perfect localization that ranks every evidence patch above every non-evidence
one. \textbf{A higher NDCG therefore means the attention scores $\mathbf{s}$ assign
larger weights to the patches that actually support the answer,
separating them from non-evidence ones.}

\begin{figure}[t]
  \centering
  \includegraphics[width=\linewidth]{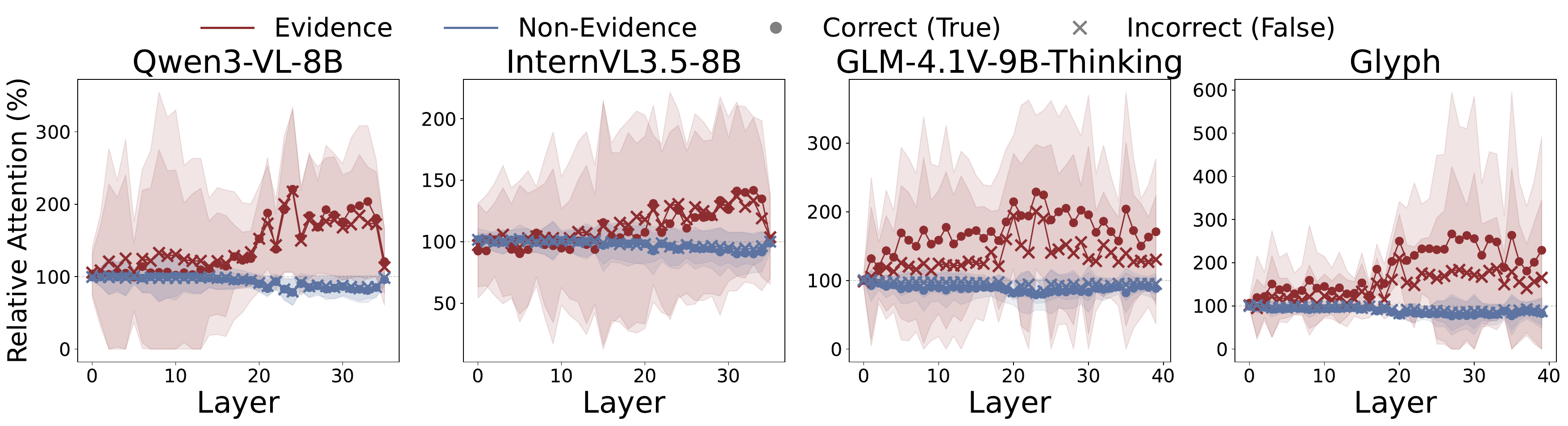}
  \caption{Relative attention (\%) per layer on HotpotQA.
  Red/blue lines: attention on Evidence/Non-Evidence tokens.
  $\bullet$/$\times$: correctly/incorrectly answered samples.}
  \label{fig:relative_attention_panel}
\end{figure}

\begin{figure}[t]
  \centering
  \includegraphics[width=\linewidth]{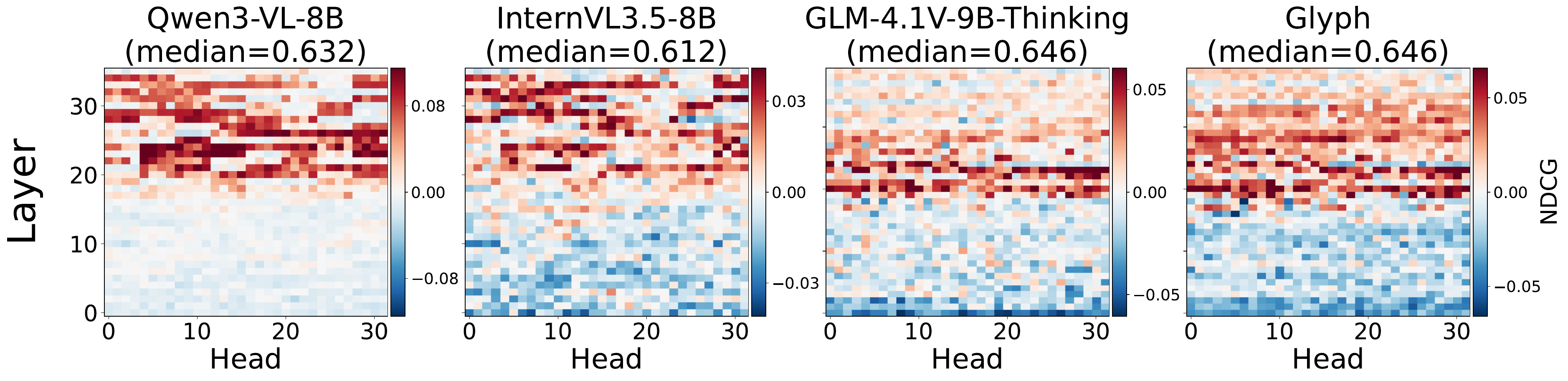}
  \caption{Per-head NDCG of last-token attention to evidence tokens on
  HotpotQA. Colors encode deviation from the per-model median NDCG.}
  \label{fig:per_head_ndcg_heatmap}
\end{figure}

\subsection{RQ1: In the VTC setting, where does evidence-localizing attention emerge inside a VLM?}
\label{sec:prelim:obs1}

To investigate \emph{where} a VLM concentrates its visual attention when
reading rendered text, we measure two complementary signals on
HotpotQA~\citep{yang2018hotpotqa} (results on other datasets are
deferred to App.~\ref{app:relative_attention_appendix}
and~\ref{app:per_head_ndcg_appendix}): (i) the average \emph{relative attention}
(\%) on evidence vs.\ non-evidence image tokens at each decoder layer
(Fig.~\ref{fig:relative_attention_panel}), where 100\% denotes the
layer-wise mean attention so values above 100\% indicate above-average
focus; and (ii) the per-head localization quality $\mathrm{NDCG}$
defined in \S\ref{sec:prelim:problem}, visualized as a
layer$\times$head heatmap (Fig.~\ref{fig:per_head_ndcg_heatmap}).

\textbf{Layer-wise emergence of evidence focus.} Across all four model
backbones, the curves in Fig.~\ref{fig:relative_attention_panel} share a
common pattern. In the early
layers (roughly the first $\sim$\,$L/2$), the relative attention on evidence
and non-evidence tokens is statistically indistinguishable, both hovering
around the layer mean of 100\%. Starting from the middle layers, evidence
attention rises sharply, opens a substantial gap above non-evidence
attention, and \emph{remains elevated through the last layer}. The shift is
consistent regardless of architecture (Qwen, InternVL, GLM-4.1V) or
post-training (GLM vs.\ Glyph). A plausible explanation is that the
early layers are still performing generic visual-feature aggregation (e.g.\
character/glyph recognition over the rendered page), while the late-middle
layers begin task-conditioned reading: once the question token has had
sufficient depth to build up a query, the model selectively routes
additional attention to the spans that actually answer it. The same trend
also appears on other datasets
(Appendix~\ref{app:relative_attention_appendix},
Fig.~\ref{fig:relative_attention_panel_appendix}).

\textbf{Per-head structure of localization.} Fig.~\ref{fig:per_head_ndcg_heatmap} shows that heads in the lower
half of the network have very limited localization ability, while the
upper half contains a population of heads whose NDCG sits well above
the median. Evidence localization is thus not a few isolated heads at random depths, but a \emph{population} that activates together at middle-to-later layers. The same pattern appears on other datasets
(App.~\ref{app:per_head_ndcg_appendix},
Fig.~\ref{fig:per_head_ndcg_heatmap_appendix}).

Together, these observations identify the middle-to-late layers as a
contiguous band where evidence-localization information is concentrated,
which helps us decide to use the localization signal from the upper
half of layers in Section~\ref{sec:methods}.\footnote{See
Fig.~\ref{fig:per_head_rank_curve_panel} for per-head NDCG vs.\ head
rank and Table~\ref{tab:layer_range_ndcg} for layer-range aggregated
NDCG.}

\begin{observation}[label={obs:1}]
Evidence-focused attention emerges at middle-to-late layers, populated by
a band of evidence-attending heads, consistently across the four VLMs we
examine.
\end{observation}

\subsection{RQ2: Are VTC failures explained by failed evidence localization?}
\label{sec:prelim:obs2}

A natural follow-up question is whether VTC failures are caused by
localization failures: when the model answers incorrectly, has it simply
failed to look at the right evidence? Fig.~\ref{fig:relative_attention_panel} splits the per-layer
relative attention curves by per-sample correctness
(SubEM\,$=1$ vs.\ SubEM\,$=0$\footnote{SubEM is a proxy for evidence
use: if evidence text appears in the answer, it likely was used.
\S\ref{sec:experiments} adopts standard EM/F1.}); circles mark
correct samples, crosses mark incorrect. Strikingly, the two subsets share the same overall trajectory. Even on
incorrectly answered samples, the relative attention on evidence tokens
rises sharply in the middle-to-late layers and stays well above the
non-evidence baseline through the last layer. A modest gap between the Correct and Incorrect curves does exist, most
visible on the GLM-family models (GLM-4.1V-9B-Thinking and Glyph),
where Incorrect samples sit slightly lower. Even so, on Incorrect samples the attention on evidence tokens
remains obviously higher than on non-evidence tokens. In other words, the model is already
\emph{looking at} the right evidence on a substantial fraction of the
samples it ultimately gets wrong. This points to a \emph{localization-without-utilization} regime: the model
correctly identifies which patches carry the evidence, yet does not always
translate that localized signal into the correct answer. Localization quality is therefore a necessary but not a sufficient
condition for VTC accuracy. Interventions that target the
\emph{utilization} side, such as making the localized evidence easier
for the model to consume, have the potential to mitigate this gap (RQ3).

\begin{observation}[label={obs:2}]
Localization quality is largely decoupled from answer correctness: incorrect
samples attend to the right evidence almost as well as correct ones. VLMs in
VTC are in a \emph{localization-without-utilization} regime.
\end{observation}

\subsection{RQ3: How can we help the model better utilize the evidence it already localizes?}
\label{sec:prelim:obs3}

\begin{wraptable}{r}{0.46\linewidth}
  \vspace{-1.0em}
  \centering
  \footnotesize
  \setlength{\tabcolsep}{4pt}
  \renewcommand{\arraystretch}{1.15}
  \caption{SubEM transition matrix on HotpotQA (Qwen3-VL-8B, 1000 samples).}
  \label{tab:gt_enlarge_subem_confusion}
  \begin{tabular}{@{}lcc@{}}
    \toprule
    & \textbf{Enlarged: T} & \textbf{Enlarged: F} \\
    \midrule
    \textbf{Plain: T} & 26.0\% (TT) & 5.5\% (TF) \\
    \textbf{Plain: F} & 21.9\% (FT) & 46.6\% (FF) \\
    \bottomrule
  \end{tabular}
  \vspace{-0.5em}
\end{wraptable}

Following the question raised by RQ2 (how to help the model better
\emph{utilize} the evidence it already attends to), we draw
inspiration from how a human reader handles a long passage. Once the
salient sentences are spotted, a natural reflex is to \emph{look
closer} at them, i.e., to magnify the relevant text. We ask whether the same intervention helps a VLM. To
isolate the effect from any
localization error, we use the ground-truth spans $\mathcal{E}$ as an oracle
and re-render each page with characters in $\mathcal{E}$ drawn at $1.5\times$
the baseline font size, leaving the rest of the page (layout, surrounding
distractors, prompt, and model weights) unchanged. Re-querying Qwen3-VL-8B
on $1{,}000$ random HotpotQA samples
(Table~\ref{tab:gt_enlarge_subem_confusion}), we observe that $21.9\%$ of
originally incorrect samples are recovered
($\mathrm{F}\!\rightarrow\!\mathrm{T}$) while only $5.5\%$ of correct ones
are broken ($\mathrm{T}\!\rightarrow\!\mathrm{F}$), giving a fix:break
ratio of $\approx 4{:}1$ and a $\sim$\,$16$-point SubEM gain---all from
changing only the on-page typography. Magnifying the important
information thus has clear potential to help the model better utilize
the evidence it already localizes.

\begin{observation}[label={obs:3}]
Localized magnification substantially recovers VTC failures, indicating its
potential to mitigate the \emph{localization-without-utilization} regime
identified in Obs.~\ref{obs:2}.
\end{observation}

\section{AGAR: Attention-Guided Adaptive Rendering}
\label{sec:methods}

\begin{wrapfigure}{r}{0.5\textwidth}
  \centering
  \includegraphics[width=0.48\textwidth]{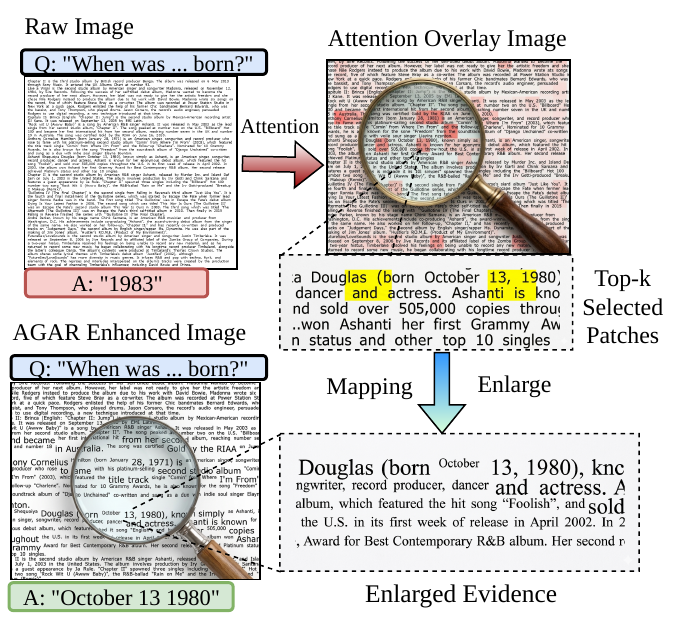}
  \caption{AGAR overview. Attention localizes evidence patches, which
    are magnified on a re-rendered page for a second forward pass.}
  \label{fig:agar}
  \vspace{-1.2em}
\end{wrapfigure}

The three observations in \S\ref{sec:preliminary} suggest a simple but
effective recipe. Obs.~\ref{obs:1} indicates \emph{where} to read the
model's own localization signal: in the middle-to-late attention
layers. Obs.~\ref{obs:2} suggests that this localization is, on its
own,  often not enough to flip a failed sample. Obs.~\ref{obs:3} shows
that physically magnifying the localized text can recover a
substantial fraction of those failures. The
natural method is then to chain the three: \emph{use the model's own attention (especially the middle-to-late
part) to decide which words to magnify, re-render the page accordingly,
and re-query the model}. We call this method \textbf{AGAR}
(\underline{A}ttention-\underline{G}uided \underline{A}daptive
\underline{R}endering); a high-level sketch is given in
Figure~\ref{fig:agar} and Algorithm~\ref{alg:agar}. AGAR proceeds in two stages, all without modifying the model weights
or the prompt. We follow the notation of \S\ref{sec:prelim:problem}.

\begin{algorithm}[t]
  \small
  \caption{AGAR: Attention-Guided Adaptive Rendering for VTC.}
  \label{alg:agar}
  \begin{algorithmic}[1]
    \Require context $T$, question $q$, VLM $\mathcal{M}$, renderer
      $\mathcal{R}$, layer range $[a, b]$, top-$K$ size $K$, font scale
      $s_{\mathrm{font}}$.
    \State $(\mathcal{I}^{(0)}, \mathcal{W}^{(0)}) \gets \mathcal{R}(T)$
      \label{alg:agar:render}
      \Comment{baseline render with word boxes}
    \State $\{\boldsymbol{\alpha}^{(\ell, h)}\} \gets \mathcal{M}\!\cdot\!\textsc{Forward}(\mathcal{I}^{(0)}, q;\, \texttt{output\_attentions})$
      \label{alg:agar:forward}
    \State $\mathbf{s} \gets \tilde{\boldsymbol{\alpha}}_{[a,b]}$
      \label{alg:agar:aggregate}
      \Comment{Stage~1: late-layer head-mean aggregation}
    \State $\mathcal{P}^{\star} \gets \mathrm{TopK}(\mathbf{s}; K)$;\;
           $\hat{\mathcal{E}} \gets \textsc{Map}(\mathcal{P}^{\star}, \mathcal{W}^{(0)})$
      \label{alg:agar:topk}
      \Comment{select top-$K$ patches; lift to char spans}
    \State $\mathcal{I}^{(1)} \gets \mathcal{R}(T,\, \hat{\mathcal{E}},\, s_{\mathrm{font}})$
      \label{alg:agar:rerender}
      \Comment{Stage~2: re-render with $\hat{\mathcal{E}}$ enlarged}
    \State $\hat{a} \gets \mathcal{M}(\mathcal{I}^{(1)}, q)$
      \label{alg:agar:answer}
      \Comment{answer from the magnified page}
    \State \Return $\hat{a}$
  \end{algorithmic}
\end{algorithm}

\paragraph{Stage 1 -- Attention-Based Evidence Localization.}
We first render the textual context $T$ at the baseline font size with
the renderer $\mathcal{R}$, obtaining the raw image
$\mathcal{I}^{(0)}$ together with a word--bounding-box map
$\mathcal{W}^{(0)} = \{(w, \beta_w, [c_w^{\,s}, c_w^{\,e}))\}$
(Alg.~\ref{alg:agar}, line~\ref{alg:agar:render}) that records, for
every rendered word $w$, its image-space box $\beta_w$ and the
character span $[c_w^{\,s}, c_w^{\,e})$ it occupies in $T$. A single
forward pass of the VLM on $(\mathcal{I}^{(0)}, q)$ with
$\texttt{output\_attentions=True}$ yields the layer-wise attentions
(Alg.~\ref{alg:agar}, line~\ref{alg:agar:forward}), which we collapse
into a per-patch score
$\mathbf{s} = \tilde{\boldsymbol{\alpha}}_{[a,b]} \in \mathbb{R}^N$ with
$[a,b] = [0.5, 1.0]$ by default (Alg.~\ref{alg:agar},
line~\ref{alg:agar:aggregate}; Obs.~\ref{obs:1}). The top-$K$ patches
$\mathcal{P}^{\star} = \mathrm{TopK}(\mathbf{s}; K)$ live in the model's
visual grid; since the unit we can actually magnify is a rendered word,
we \emph{map} the selection from patches to character spans:
$\hat{\mathcal{E}} = \textsc{Map}(\mathcal{P}^{\star}, \mathcal{W}^{(0)})$
(Alg.~\ref{alg:agar}, line~\ref{alg:agar:topk}).
Concretely, $\textsc{Map}$ marks a word $w \in \mathcal{W}^{(0)}$ as
selected whenever its box $\beta_w$ overlaps any selected patch
$p_i \in \mathcal{P}^{\star}$ (after rescaling $\beta_w$ to the visual
encoder's input resolution), collects the corresponding character spans
$[c_w^{\,s}, c_w^{\,e})$, and merges overlapping/adjacent spans.
The output $\hat{\mathcal{E}} \subseteq T$ is our \emph{predicted}
evidence---the attention-only counterpart of the oracle $\mathcal{E}$ used
in Obs.~\ref{obs:3}.

\paragraph{Stage 2 -- Magnified Re-rendering and Re-inference.}
We re-invoke the renderer with $\hat{\mathcal{E}}$ as the set of character
spans to draw at a magnified font size $s_{\mathrm{font}}$,
$\mathcal{I}^{(1)} = \mathcal{R}(T,\, \hat{\mathcal{E}},\, s_{\mathrm{font}})$
(Alg.~\ref{alg:agar}, line~\ref{alg:agar:rerender}), while keeping all
other typography and layout parameters identical to $\mathcal{I}^{(0)}$. We then produce the
final answer with one more forward pass,
$\hat{a} = \mathcal{M}(\mathcal{I}^{(1)}, q)$ (Alg.~\ref{alg:agar},
line~\ref{alg:agar:answer}), and discard the first-pass output. This makes AGAR a model-agnostic plug-in for any VLM
that exposes layer-wise attentions. No prompt or weight changes are
needed.

\newcommand{\up}[1]{\,{\scriptsize\textcolor{teal!75!black}{(+#1\%)}}}
\newcommand{\dn}[1]{\,{\scriptsize\textcolor{red!70!black}{($-$#1\%)}}}
\newcommand{\eq}{\,{\scriptsize\textcolor{gray}{($\pm$0\%)}}}

\section{Experiments}
\label{sec:experiments}

We evaluate AGAR around four questions:
(i)~does AGAR consistently improve VTC accuracy on off-the-shelf VLMs
across short-form, long-context, and multi-page memory QA
(\S\ref{sec:exp:main})? (ii)~does its gain persist when the underlying VLM
is post-trained (\S\ref{sec:exp:sft})? (iii)~is the gain robust
to low-quality input images and irrelevant text contexts
(\S\ref{sec:exp:robustness})? and (iv)~how sensitive is AGAR to its two
core hyper-parameters, the patch budget $K$ and the magnification factor
$s_{\mathrm{font}}$ (\S\ref{sec:exp:analysis})?

\subsection{Experimental Setup}
\label{sec:exp:setup}

\paragraph{Datasets.}
We benchmark AGAR on three families of VTC tasks of increasing visual and
textual complexity. (i)~\textbf{Short-form QA}: NQ~\citep{kwiatkowski-etal-2019-natural},
HotpotQA~\citep{yang2018hotpotqa} (distractor mode),
TriviaQA~\citep{joshi2017triviaqa} (TQA), and
NewsQA~\citep{trischler2017newsqa}, each rendered onto a single page.
(ii)~\textbf{LongBench~\citep{bai2024longbench} multi-page QA}: HotpotQA-LB
(HQA), 2WikiMultihop~\citep{ho2020constructing} (2WMQA),
MuSiQue~\citep{trivedi2022musique} (MSQ), and
Qasper~\citep{dasigi2021dataset} (QP), where the context spans multiple
rendered pages. (iii)~\textbf{LoCoMo~\citep{maharana2024evaluating}
multi-page memory QA}, which further stresses long, session-style
contexts. We use the official
validation/test splits and report EM and F1; per-dataset sample counts and
average context lengths are summarized in Table~\ref{tab:dataset_stats}
(Appendix~\ref{app:dataset_stats}).

\paragraph{Models and Baselines.}
Our primary backbone is \textbf{Qwen3-VL-8B-Instruct}; cross-architecture
results on \textbf{InternVL3.5-8B}, \textbf{GLM-4.1V-9B-Thinking}, and
\textbf{Glyph} (post-trained on GLM-4.1V-9B-Thinking) are in
Appendix~\ref{app:cross_arch}. We compare AGAR against three baseline groups:
\emph{Plain VQA} (answers directly from $\mathcal{I}^{(0)}$); two
\emph{naive baselines} that isolate AGAR's two design choices ---
\emph{random}, which magnifies $K$ randomly chosen patches (tests
\emph{which} patches matter), and \emph{uniform}, which scales the whole
page font to match AGAR's visual-token count (tests whether token count
alone suffices); and prior emphasis methods adapted from VQA:
\emph{VEA}~\citep{liu2025seeing}, \emph{CGR}/\emph{VAR}~\citep{liu2025unveiling},
and \emph{AGLA}~\citep{an2025mitigating} (App.~\ref{app:baseline_impl}).

\paragraph{Implementation.}
Pages render at $9$\,px; AGAR uses late-layer range $[a,b]=[0.5,1.0]$ and
font scale $s_{\mathrm{font}}=1.5$. The patch budget is $K=8$ for
single-page short-form QA and $K\!\approx\!2\%$ of total patches for the
multi-page benchmarks (LongBench, LoCoMo). See examples in App.~\ref{app:case_study}.

\subsection{Main Results: AGAR on Off-the-Shelf VLMs}
\label{sec:exp:main}

To answer question~(i), we evaluate off-the-shelf Qwen3-VL-8B-Instruct
on the abovementioned datasets. Short-form single-page QA results are
reported in Tab.~\ref{tab:exp_main_shortform}. Multi-page LongBench QA
and LoCoMo memory QA results are reported jointly in
Tab.~\ref{tab:exp_main_multipage}. The two naive baselines (Random, Uniform) yield no or only marginal
gains over Plain VQA. This indicates that AGAR's gain comes from
effectively selecting the important evidence to emphasize, rather than
simply increasing the number of visual tokens. The prior VQA-emphasis methods (VEA, CGR, VAR, AGLA) provide only
limited enhancement and underperform our method, and even hurt the
performance on some datasets. These methods are originally designed
for standard VQA, where \emph{de-emphasising} irrelevant visual
background helps the answer. They thus import the wrong inductive bias
for VTC, in which the surrounding text must remain readable for full
context understanding.
AGAR, in contrast, delivers clear and consistent gains over Plain VQA
on every cell of both tables (e.g.\ $+18.1\%$ F1 on short-form
HotpotQA, $+28.5\%$ on LB-MSQ multi-page, $+38.8\%$ on LoCoMo
OpenDomain) and strictly outperforms every baseline, while increasing
visual-token cost only marginally (maintaining
$\sim$\,$3\times$ compression). The same observation holds for the
other three VLMs; see Appendix~\ref{app:cross_arch},
Tables~\ref{tab:cross_arch_internvl_short}--\ref{tab:cross_arch_glyph_multi}.
\begin{table}[ht]
  \centering
  \footnotesize
  \setlength{\tabcolsep}{3pt}
  \renewcommand{\arraystretch}{1.1}
  \caption{\textbf{Short-form single-page QA} on Qwen3-VL-8B. Each cell shows
  the metric followed by its \emph{relative} change (\%) vs.\ Plain VQA.
  Avg.\ text tokens $\approx 805$ ($\sim$\,$3\times$ compression).}
  \label{tab:exp_main_shortform}
  \resizebox{\linewidth}{!}{%
  \begin{tabular}{@{}l cc cc cc cc | r@{}}
    \toprule
    & \multicolumn{2}{c}{\textbf{NQ}} & \multicolumn{2}{c}{\textbf{HotpotQA}}
    & \multicolumn{2}{c}{\textbf{NewsQA}} & \multicolumn{2}{c|}{\textbf{TQA}}
    & \textbf{Avg.\ tok.} \\
    \cmidrule(lr){2-3}\cmidrule(lr){4-5}\cmidrule(lr){6-7}\cmidrule(lr){8-9}
    \textbf{Method} & EM & F1 & EM & F1 & EM & F1 & EM & F1 & ($\downarrow$) \\
    \midrule
    Plain VQA & 47.3 & 63.8 & 29.2 & 39.7 & 36.7 & 58.3 & 61.4 & 67.9 & 268 \\
    Random    & 47.1\dn{0.4} & 63.7\dn{0.2} & 30.4\up{4.1} & 41.8\up{5.3} & 37.6\up{2.5} & 58.7\up{0.7} & 63.2\up{2.9} & 70.4\up{3.7} & 301 \\
    Uniform   & 48.2\up{1.9} & 64.4\up{0.9} & 30.3\up{3.8} & 41.4\up{4.3} & 37.6\up{2.5} & 58.7\up{0.7} & 62.6\up{2.0} & 69.6\up{2.5} & 292 \\
    VEA-Br    & 47.2\dn{0.2} & 63.9\up{0.2} & 30.0\up{2.7} & 40.5\up{2.0} & 36.2\dn{1.4} & 57.7\dn{1.0} & 60.7\dn{1.1} & 68.6\up{1.0}  & 268 \\
    VEA-Con   & 47.2\dn{0.2} & 63.9\up{0.2} & 30.9\up{5.8} & 41.6\up{4.8} & 36.4\dn{0.8} & 58.1\dn{0.3} & 59.9\dn{2.4} & 67.8\dn{0.1}  & 268 \\
    CGR       & 11.5\dn{75.7} & 29.3\dn{54.1} & 22.6\dn{22.6} & 29.5\dn{25.7} & 38.5\up{4.9} & 59.3\up{1.7} & 51.5\dn{16.1} & 67.3\dn{0.9} & 268 \\
    VAR       & 46.9\dn{0.8}  & 63.6\dn{0.3}  & 26.8\dn{8.2}  & 38.3\dn{3.5}  & 13.5\dn{63.2} & 30.9\dn{47.0} & 46.9\dn{23.6} & 63.6\dn{6.3} & 268 \\
    AGLA      & 48.8\up{3.2}  & 64.6\up{1.3}  & 30.2\up{3.4}  & 41.2\up{3.8}  & 37.2\up{1.4} & 58.7\up{0.7} & 63.2\up{2.9}  & 70.2\up{3.4} & 268 \\
    \midrule
    \textbf{AGAR (ours)} & \textbf{50.2}\up{6.1} & \textbf{66.5}\up{4.2} & \textbf{34.3}\up{17.5} & \textbf{46.9}\up{18.1} & \textbf{39.0}\up{6.3} & \textbf{60.3}\up{3.4} & \textbf{64.7}\up{5.4} & \textbf{72.2}\up{6.3} & 291 \\
    \bottomrule
  \end{tabular}%
  }
\end{table}

\begin{table}[ht]
  \centering
  \footnotesize
  \setlength{\tabcolsep}{3pt}
  \renewcommand{\arraystretch}{1.1}
  \caption{\textbf{Multi-page benchmarks} on Qwen3-VL-8B (F1). \emph{Left}:
  LongBench multi-page QA (LB-HQA / LB-2WMQA / LB-MSQ / LB-QP). \emph{Right}:
  LoCoMo multi-page memory QA (SingleHop / MultiHop / Temporal / OpenDomain).
  Each cell shows F1 followed by its \emph{relative} change (\%) vs.\
  Plain VQA. Avg.\ text
  tokens $\approx 14.7$k ($\sim$\,$3\times$ compression).}
  \label{tab:exp_main_multipage}
  \resizebox{\linewidth}{!}{%
  \begin{tabular}{@{}l cccc | cccc | r@{}}
    \toprule
    & \multicolumn{4}{c|}{\textbf{LongBench (multi-page) -- F1}}
    & \multicolumn{4}{c|}{\textbf{LoCoMo (multi-page memory) -- F1}}
    & \textbf{Avg.\ tok.} \\
    \cmidrule(lr){2-5}\cmidrule(lr){6-9}
    \textbf{Method}
      & \textbf{LB-HQA} & \textbf{LB-2WMQA} & \textbf{LB-MSQ} & \textbf{LB-QP}
      & \textbf{SingleHop} & \textbf{MultiHop} & \textbf{Temporal} & \textbf{OpenDom.}
      & ($\downarrow$) \\
    \midrule
    Plain VQA & 39.8            & 37.5            & 17.2            & 29.5
              & 26.1            & 20.5            & 22.7            & 12.1
              & 4{,}916 \\
    Random    & 40.8\up{2.5}    & 37.3\dn{0.5}    & 17.6\up{2.3}    & 33.8\up{14.6}
              & 28.0\up{7.3}    & 19.0\dn{7.3}    & 22.6\dn{0.4}    & 14.8\up{22.3}
              & 5{,}351 \\
    Uniform   & 39.8\eq         & 37.5\eq         & 16.9\dn{1.7}    & 33.3\up{12.9}
              & 26.5\up{1.5}    & 18.5\dn{9.8}    & 21.8\dn{4.0}    & 12.2\up{0.8}
              & 5{,}378 \\
    VEA-Br    & 41.8\up{5.0}    & 35.2\dn{6.1}    & 21.3\up{23.8}   & 30.3\up{2.7}
              & 29.4\up{12.6}   & 20.3\dn{1.0}    & 23.5\up{3.5}    & 15.4\up{27.3}
              & 4{,}916 \\
    VEA-Con   & 42.1\up{5.8}    & 36.1\dn{3.7}    & 18.2\up{5.8}    & 29.9\up{1.4}
              & 27.8\up{6.5}    & 19.2\dn{6.3}    & 23.1\up{1.8}    & 13.5\up{11.6}
              & 4{,}916 \\
    CGR       & 34.1\dn{14.3}   & 36.6\dn{2.4}    & 21.2\up{23.3}   & 29.3\dn{0.7}
              & 30.1\up{15.3}   & 21.8\up{6.3}    & 21.2\dn{6.6}    & 15.8\up{30.6}
              & 4{,}916 \\
    VAR       & 37.7\dn{5.3}    & 39.2\up{4.5}    & 18.5\up{7.6}    & 30.9\up{4.7}
              & 24.3\dn{6.9}    & 17.1\dn{16.6}   & 20.0\dn{11.9}   & 16.0\up{32.2}
              & 4{,}916 \\
    AGLA      & 41.2\up{3.5}    & 39.1\up{4.3}    & 21.7\up{26.2}   & 31.5\up{6.8}
              & 26.1\eq         & 19.1\dn{6.8}    & 24.8\up{9.3}    & 15.2\up{25.6}
              & 4{,}916 \\
    \midrule
    \textbf{AGAR (ours)} & \textbf{44.3}\up{11.3} & \textbf{40.8}\up{8.8} & \textbf{22.1}\up{28.5} & \textbf{34.2}\up{15.9}
              & \textbf{32.0}\up{22.6} & \textbf{21.9}\up{6.8} & \textbf{25.7}\up{13.2} & \textbf{16.8}\up{38.8}
              & 5{,}384 \\
    \bottomrule
  \end{tabular}%
  }
\end{table}

\begin{takeaway}
AGAR effectively improves VTC performance on off-the-shelf VLMs with
only a marginal increase in visual-token cost.
\end{takeaway}

\subsection{Composing AGAR with Post-Training}
\label{sec:exp:sft}

\begin{wrapfigure}[14]{r}{0.45\linewidth}
  \centering
  \includegraphics[width=\linewidth]{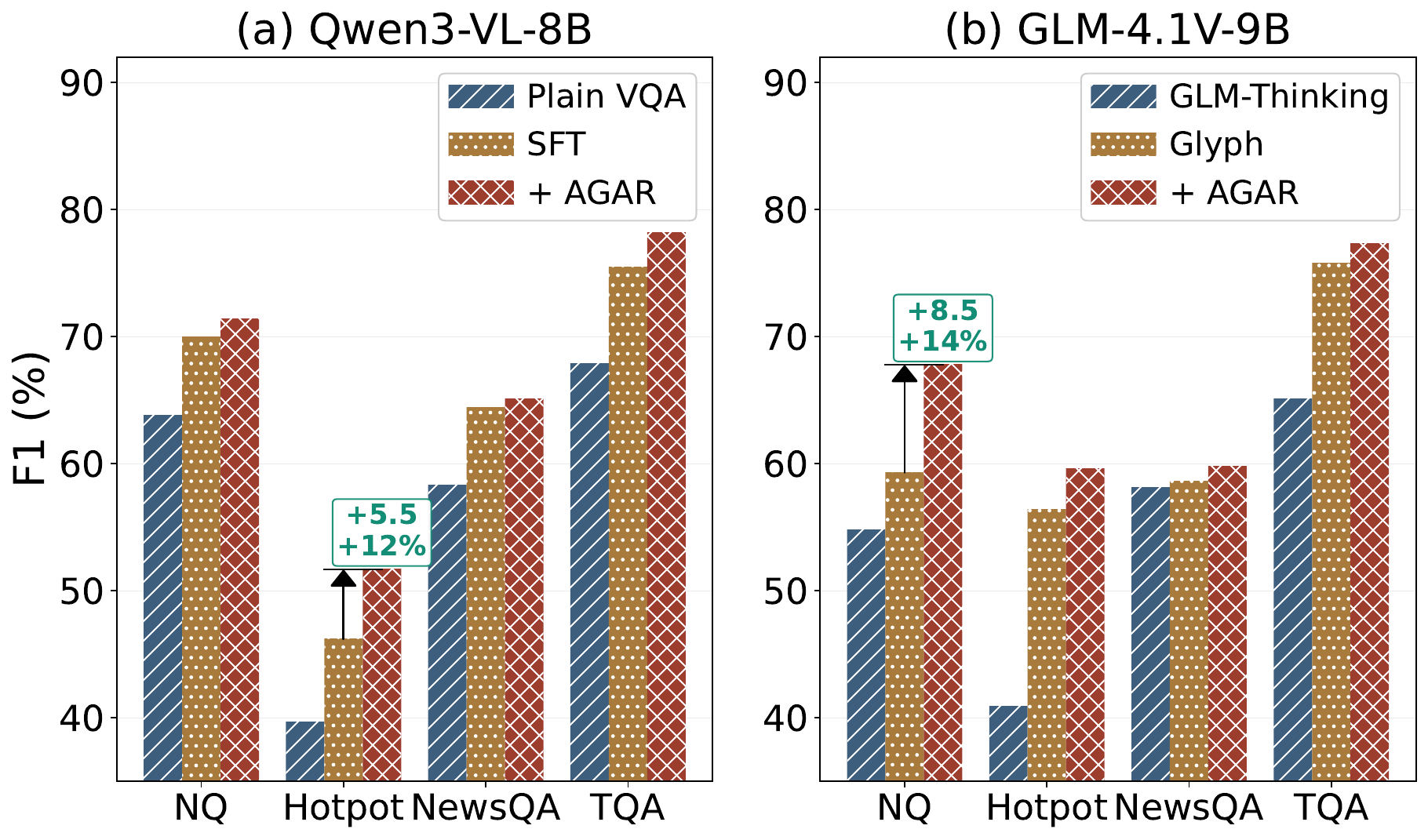}
  \caption{AGAR composes with post-training (short-form F1).
\textbf{(a)} Qwen3-VL-8B vs.\ mix4 SFT.
\textbf{(b)} GLM-4.1V-9B-Thinking vs.\ Glyph.}
  \label{fig:exp_sft}
\end{wrapfigure}

Many real-world VTC applications~\citep{cheng2025glyph,
shi2026memocr, wang2026vtc} adapt VLMs via post-training, raising
the question of whether AGAR's gain persists on post-trained VLMs.
We test this on the short-form QA family (NQ / HotpotQA / NewsQA /
TQA) under two settings. \emph{(a) Qwen3-VL-8B-Instruct vs.\ its
SFT counterpart} (Fig.~\ref{fig:exp_sft}(a)): we LoRA-fine-tune
Qwen3-VL-8B-Instruct on \emph{mix4}, a balanced $20$k corpus sampled from the four training splits ($5$k each, rendered at the
standard $9$\,px, no attention-aware augmentation), and evaluate on
the same held-out splits as \S\ref{sec:exp:main}. \emph{(b)
GLM-4.1V-9B-Thinking vs.\ Glyph} (Fig.~\ref{fig:exp_sft}(b)):
\textbf{Glyph}~\citep{cheng2025glyph} is a public third-party VTC
model derived from GLM-4.1V-9B-Thinking via large-scale SFT+RL.

In both settings, AGAR is applied at inference only, with no change
to training data, objective, or weights. As shown in Fig.~\ref{fig:exp_sft}, AGAR continues to lift F1 on top
of every post-trained baseline across all four datasets in both
regimes. On some datasets, the further gain AGAR delivers over the
post-trained model is even comparable to, or exceeds, the gain
post-training itself delivered: on \emph{(a)}~HotpotQA, Plain
VQA~$\to$~SFT lifts F1 by $+6.5$ absolute, and SFT~$\to$~SFT~+~AGAR
adds another $+5.5$; on \emph{(b)}~NQ, GLM-Thinking~$\to$~Glyph lifts
F1 by $+4.5$ while Glyph~$\to$~Glyph~+~AGAR adds another $+8.5$ on
top. AGAR therefore
composes effectively with VLM post-training to deliver a further free
inference-time boost.

\begin{takeaway}
AGAR composes effectively with VLM post-training to further improve
performance.
\end{takeaway}
\vspace{-0.6em}

\subsection{Robustness to Imperfect Inputs}
\label{sec:exp:robustness}
\vspace{-0.4em}

\begin{figure}[ht]
  \centering
  \includegraphics[width=\linewidth]{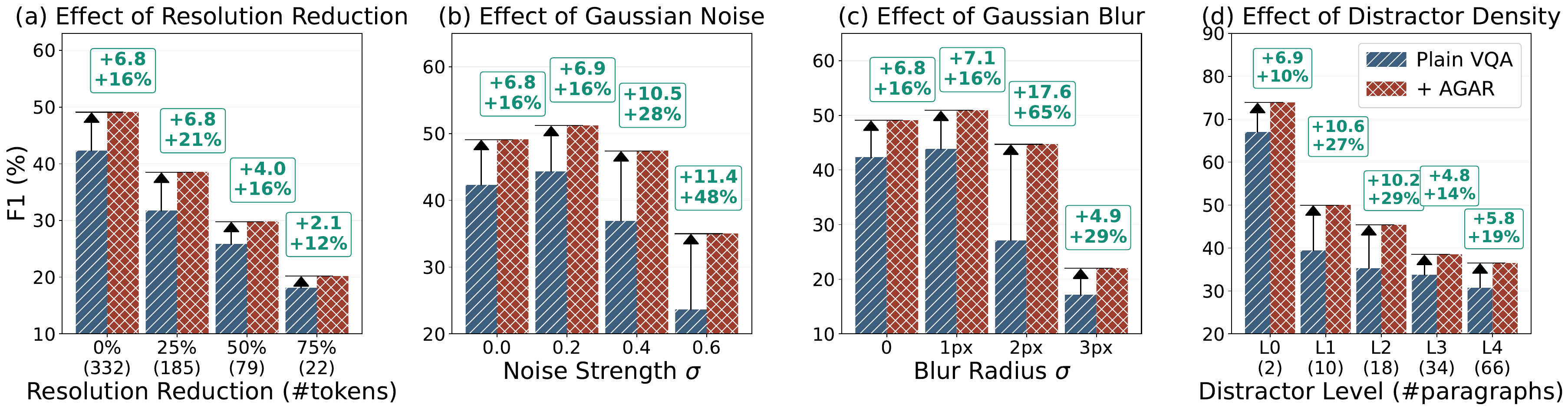}
  \caption{\textbf{AGAR robustness on Qwen3-VL-8B / HotpotQA.} Blue:
Plain F1; orange: AGAR F1; box: absolute and relative $\Delta$.
\textbf{(a)--(c)}: \textbf{visual} corruptions (downsampling,
Gaussian noise, blur). \textbf{(d)}: \textbf{text-side} dilution
from L0 (gold only) to L4 (66 paragraphs, with hard negatives at
L2--L4).}
  \label{fig:robustness_panel}
\end{figure}

We further ask whether AGAR remains useful under degraded
inputs---common in real-world VTC where pages are scanned,
compressed, or surrounded by retrieval noise. We probe two principal
degradation sources on Qwen3-VL-8B: \emph{(i)~image-level
perturbations} (low resolution, sensor noise, defocus blur) and
\emph{(ii)~text-side distractors} surrounding the gold evidence
(e.g., unrelated paragraphs from retrieval). Fig.~\ref{fig:robustness_panel}
reports plain F1, AGAR F1, and the gain $\Delta = \mathrm{F1_{enh}}
- \mathrm{F1_{plain}}$.

\vspace{-0.6em}

\paragraph{Visual perturbations.}
We perturb the rendered HotpotQA pages (1000 samples) with three families of
image corruption applied directly to the PNG before inference: resolution
reduction (downsample by $\{25\%, 50\%, 75\%\}$, which simultaneously
shrinks the visual-token budget from $332$ to $\{185, 79, 22\}$ tokens),
additive Gaussian noise on pixel intensities ($\sigma\in\{0.2, 0.4, 0.6\}$
on the $[0,1]$ range), and Gaussian blur ($\sigma\in\{1,2,3\}$ pixels);
plain vs.\ AGAR inference is then run on the perturbed images
(Fig.~\ref{fig:robustness_panel}~(a)--(c)). The main observation is that AGAR is robust to visual degradation and
recovers a large fraction of the F1 that Plain VQA loses under it. Under
resolution reduction, for example, AGAR at $79$ visual tokens ($50\%$
downsample) matches Plain VQA at $185$ tokens ($25\%$ downsample) ---
AGAR effectively buys back a full level of image quality / token budget.
Moreover, in some regimes AGAR's recovery actually \emph{grows} with the
severity of the perturbation: blur $\sigma{=}2$\,px yields
$\Delta{=}+17.6$ ($+65\%$) and Gaussian noise $\sigma{=}0.6$ yields
$\Delta{=}+11.4$ ($+48\%$), against an unperturbed baseline of
$\Delta{=}+6.8$ ($+16\%$).

\paragraph{Text-side irrelevance.}
We build a $5$-level distractor-density ladder on HotpotQA (1000 samples
per level) mimicking a retrieval pipeline of increasing noise: L0 = $2$
gold only; L1 = native HotpotQA distractor ($2$ gold $+$ $8$
distractors); L2--L4 add $\{8, 24, 56\}$ hard negatives drawn from other
HotpotQA training questions. Gold paragraphs are always present but
randomly positioned. Plain F1 cliffs at L0$\to$L1 ($67.0\to 39.4$,
$-27.6$) and then drifts down only mildly ($\to 30.7$ at L4). AGAR's
$\Delta$ stays positive throughout, with the relative gain at every
distractor level exceeding the clean-input (L0) baseline, and peaks
under moderate clutter (L1: $+10.6$, $+27\%$; L2: $+10.2$, $+29\%$).

\begin{takeaway}
AGAR is robust to both visual and text-side input degradation, and
effectively recovers much of the F1 loss caused by quality degradation.
\end{takeaway}

\subsection{Hyperparameter Analysis}
\label{sec:exp:analysis}

\begin{wrapfigure}{r}{0.5\linewidth}
  \vspace{-1.0\baselineskip}
  \centering
  \includegraphics[width=\linewidth]{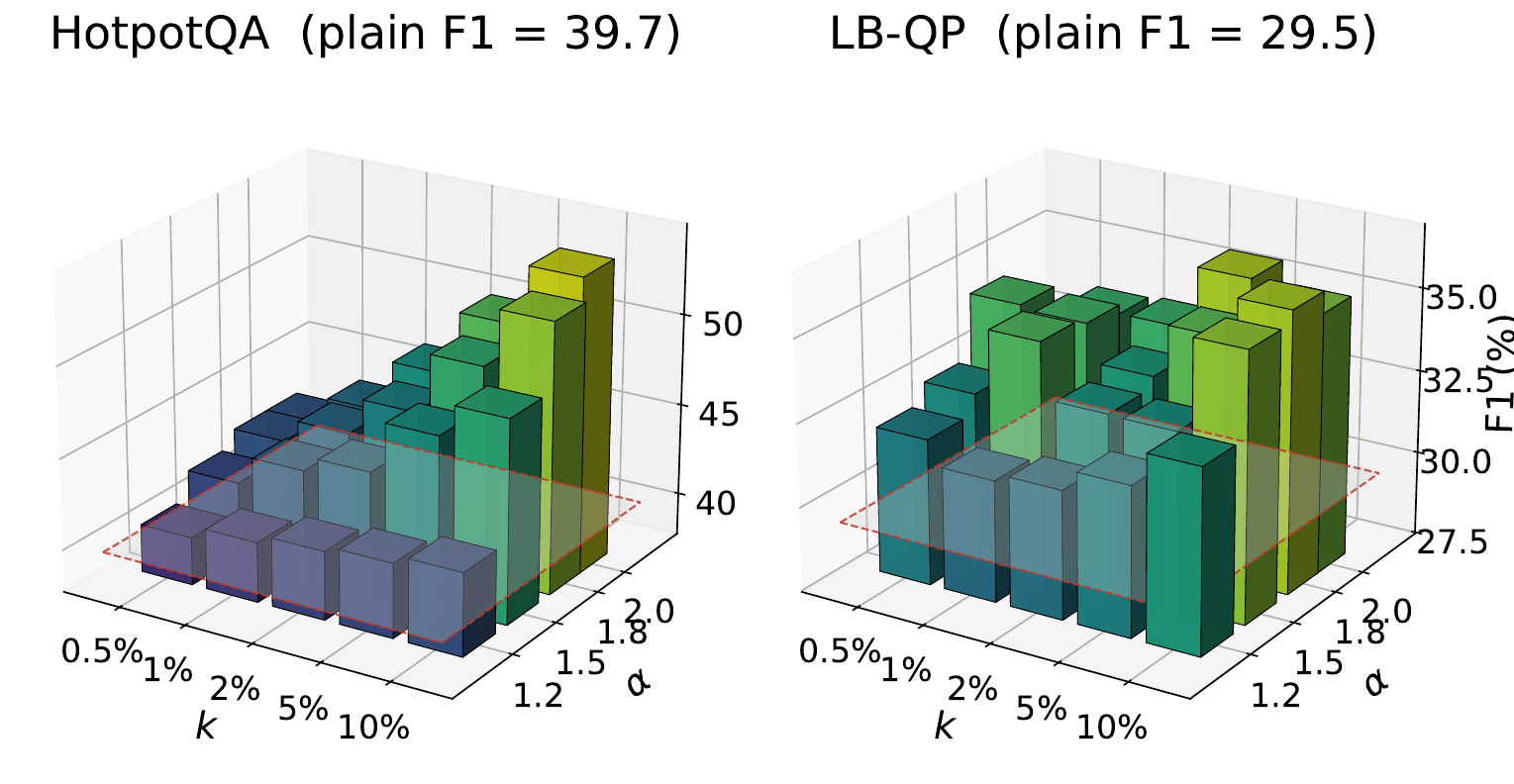}
  \caption{\textbf{AGAR's $(k,\alpha)$ landscape} on Qwen3-VL-8B on
  HotpotQA and LB-QP. The red dashed plane marks Plain VQA.}
  \label{fig:sensitivity_2col}
  \vspace{-0.6\baselineskip}
\end{wrapfigure}

To answer~(iv), we study AGAR's sensitivity to its two
hyperparameters: the fraction $k$ of patches to magnify and the
font scale $\alpha$ applied to the covered words. We sweep
$k \in \{0.5\%, 1\%, 2\%, 5\%, 10\%\}$ and
$\alpha \in \{1.2, 1.5, 1.8, 2.0\}$ on all datasets from
\S\ref{sec:exp:main}. Fig.~\ref{fig:sensitivity_2col} shows
HotpotQA (single-page) and LB-QP (multi-page) as representatives;
the full sweep is in Appendix~\ref{app:sensitivity}. AGAR is \emph{robust} to $(k,\alpha)$: nearly all cells exceed the
Plain VQA plane (HotpotQA: $20/20$, LB-QP: $19/20$; $11/12$ appendix
subtasks have $\ge 90\%$ above-plane cells). Small $(k,\alpha)$
underperform due to insufficient magnification against attention
dilution; larger values trade tokens for accuracy, tunable to budget.
Well-chosen pairs yield pronounced gains (HotpotQA F1
$+14.3$/$+35.9\%$ at $(10\%, 2.0)$). Other
hyperparameters (base font size, layer range) are discussed in
App.~\ref{app:token_base} and~\ref{app:layer_range_ndcg}.

\begin{takeaway}
AGAR is robust to its $(k,\alpha)$ hyperparameters: most pairs already
improve over Plain VQA, while stronger $(k,\alpha)$ trade additional
visual tokens for substantially larger gains.
\end{takeaway}
\vspace{-0.9em}
\section{Conclusion and Limitation}
\vspace{-0.7em}

We studied how VLMs internally process Visual Text Comprehension
(VTC) inputs and where their failures come from. VLMs reliably
localize evidence in middle-to-late layers, yet this localization is
decoupled from answer correctness, leaving them in a
\emph{localization-without-utilization} regime. Magnifying the
localized evidence already substantially repairs these failures.
Motivated by this, we propose \textbf{AGAR}, a plug-and-play method
that uses the model's own attention to enlarge identified evidence
patches and re-query the same VLM. Across four open-weight VLMs and
12 subtasks, AGAR consistently improves over Plain VQA,
outperforms training-free baselines, composes with post-training,
and stays robust under visual- and text-side degradation.\\
\textbf{Limitations.} AGAR requires access to the backbone's
attention scores, limiting its use on closed-source VLMs. Deeper
mechanistic accounts (e.g., circuit-level analysis) are left to future work.

\bibliographystyle{plainnat}
\bibliography{references}


\clearpage
\appendix

\section*{\textsc{Appendix}}
\begingroup
\setlength{\parskip}{0.25em}
\par\noindent(\S\ref{app:sensitivity})~~\nameref{app:sensitivity}\dotfill\pageref{app:sensitivity}
\par\noindent(\S\ref{app:dataset_stats})~~\nameref{app:dataset_stats}\dotfill\pageref{app:dataset_stats}
\par\noindent(\S\ref{app:cross_arch})~~\nameref{app:cross_arch}\dotfill\pageref{app:cross_arch}
\par\noindent(\S\ref{app:relative_attention_appendix})~~\nameref{app:relative_attention_appendix}\dotfill\pageref{app:relative_attention_appendix}
\par\noindent(\S\ref{app:per_head_ndcg_appendix})~~\nameref{app:per_head_ndcg_appendix}\dotfill\pageref{app:per_head_ndcg_appendix}
\par\noindent(\S\ref{app:layer_range_ndcg})~~\nameref{app:layer_range_ndcg}\dotfill\pageref{app:layer_range_ndcg}
\par\noindent(\S\ref{app:baseline_impl})~~\nameref{app:baseline_impl}\dotfill\pageref{app:baseline_impl}
\par\noindent(\S\ref{app:token_base})~~\nameref{app:token_base}\dotfill\pageref{app:token_base}
\par\noindent(\S\ref{app:case_study})~~\nameref{app:case_study}\dotfill\pageref{app:case_study}
\par\noindent(\S\ref{app:compute})~~\nameref{app:compute}\dotfill\pageref{app:compute}
\par\noindent(\S\ref{app:broader_impact})~~\nameref{app:broader_impact}\dotfill\pageref{app:broader_impact}
\par\noindent(\S\ref{app:prompt})~~\nameref{app:prompt}\dotfill\pageref{app:prompt}
\endgroup
\vspace{1em}

\section{Full $(k,\alpha)$ Sensitivity Sweep}
\label{app:sensitivity}

\begin{figure}[h]
  \centering
  \includegraphics[width=\linewidth]{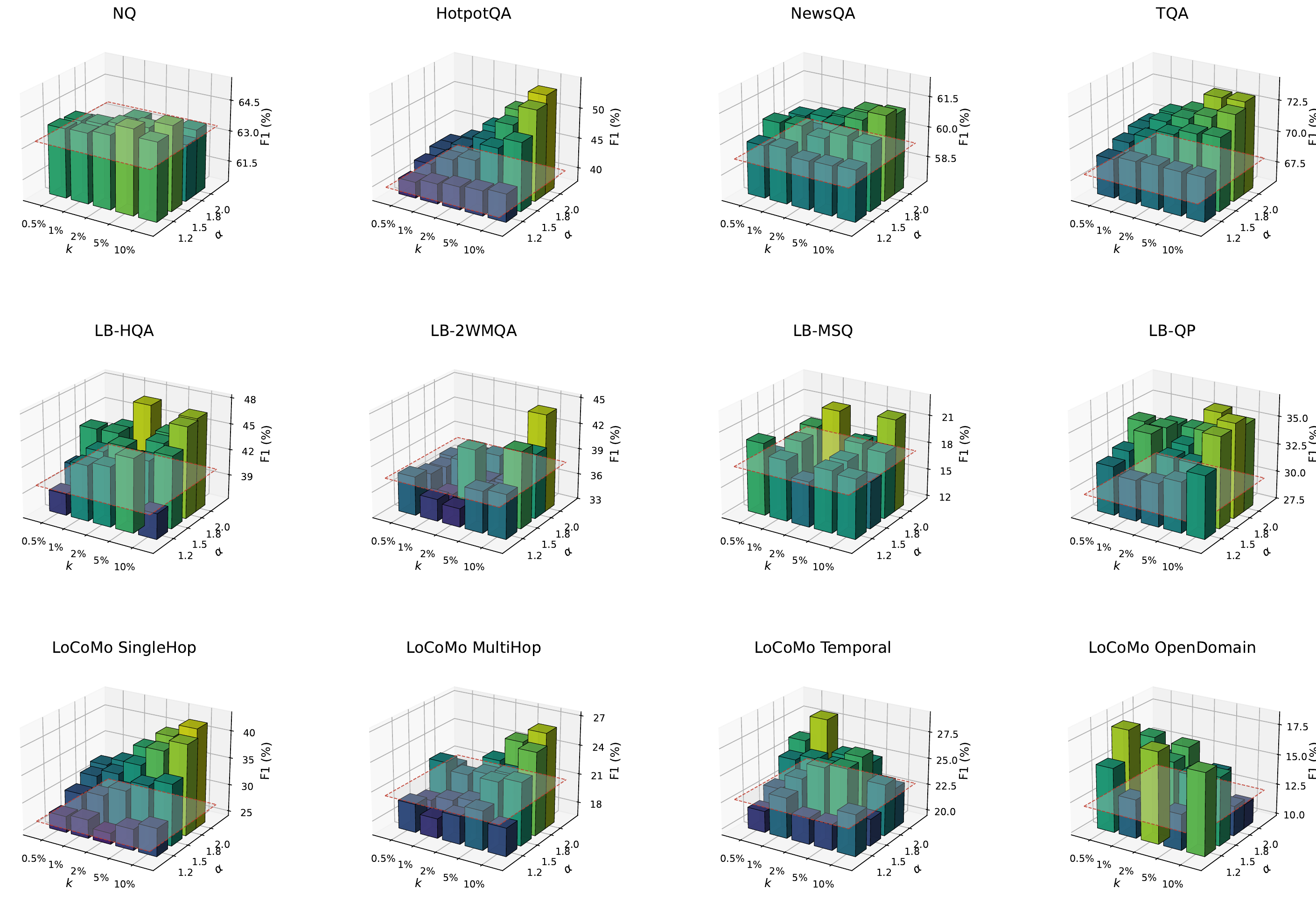}
  \caption{\textbf{Full $(k,\alpha)$ sweep on 12 subtasks.} 3D bar version.
  Short-form QA (top row): NQ / HotpotQA / NewsQA / TQA. LongBench
  multi-page (middle row): LB-HQA / LB-2WMQA / LB-MSQ / LB-QP. LoCoMo
  multi-page memory (bottom row): SingleHop / MultiHop / Temporal /
  OpenDomain.}
  \label{fig:sensitivity_full}
\end{figure}

Fig.~\ref{fig:sensitivity_full} extends the headline 1$\times$2 view of
\S\ref{sec:exp:analysis} (HotpotQA + LB-QP) to all 12 subtasks used in
\S\ref{sec:exp:main}. Each panel reports enhanced F1 (\%) at every
$(k,\alpha) \in \{0.5\%, 1\%, 2\%, 5\%, 10\%\} \times \{1.2, 1.5, 1.8, 2.0\}$
cell; the translucent grey plane (red dashed wireframe) marks the
Plain~VQA baseline, and per-panel colour is normalised to the panel's
own range.

\section{Dataset Statistics}
\label{app:dataset_stats}

Table~\ref{tab:dataset_stats} summarizes the three benchmark families used in
\S\ref{sec:experiments}: short-form QA (single rendered page), LongBench
multi-page QA, and LoCoMo multi-page memory QA. Numbers are computed on the
official validation/test split of each dataset (HotpotQA in distractor mode);
context length is reported in both raw characters and whitespace-separated
words.

\begin{table}[h]
  \centering
  \footnotesize
  \setlength{\tabcolsep}{6pt}
  \renewcommand{\arraystretch}{1.1}
  \caption{Dataset statistics for the three benchmark families used in
  \S\ref{sec:experiments}. \textbf{Avg.\ chars} and \textbf{Avg.\ words}
  are the per-sample mean character and whitespace-separated word counts
  of the textual context~$T$ before rendering. All datasets are used
  consistently with their licenses for non-commercial academic research.}
  \label{tab:dataset_stats}
  \begin{tabular}{@{}llrrrl@{}}
    \toprule
    \textbf{Family} & \textbf{Dataset} & \textbf{\# Samples}
      & \textbf{Avg.\ chars} & \textbf{Avg.\ words} & \textbf{License} \\
    \midrule
    \multirow{4}{*}{\shortstack[l]{Short-form QA\\(single page)}}
      & NQ        & 12{,}836 &     843 &    159 & CC BY-SA 3.0 \\
      & HotpotQA  &  7{,}404 &  5{,}479 &    897 & CC BY-SA 4.0 \\
      & NewsQA    &  4{,}212 &  3{,}001 &    493 & MIT \\
      & TQA       &  7{,}785 &  4{,}064 &    673 & Apache 2.0 \\
    \midrule
    \multirow{4}{*}{\shortstack[l]{LongBench\\(multi-page)}}
      & LB-HQA    &    200 &  56{,}550 &  9{,}133 & MIT \\
      & LB-2WMQA  &    200 &  29{,}615 &  4{,}873 & MIT \\
      & LB-MSQ    &    200 &  69{,}405 & 11{,}196 & MIT \\
      & LB-QP     &    200 &  23{,}641 &  3{,}599 & MIT \\
    \midrule
    Memory QA
      & LoCoMo    &  1{,}542 &  79{,}668 & 14{,}394 & CC BY-NC 4.0 \\
    \bottomrule
  \end{tabular}
\end{table}

\section{Cross-Architecture Results}
\label{app:cross_arch}

To verify that the AGAR gain is not specific to one architecture or
post-training recipe, we additionally evaluate three other VLMs:
\textbf{InternVL3.5-8B}, \textbf{GLM-4.1V-9B-Thinking}, and
\textbf{Glyph} (a public model post-trained from GLM-4.1V-9B-Thinking
specifically for visual text comprehension). Tables~\ref{tab:cross_arch_internvl_short}--\ref{tab:cross_arch_glyph_multi}
mirror Tables~\ref{tab:exp_main_shortform}--\ref{tab:exp_main_multipage}
of the main text. The magnitude and direction of the gain track
the same patterns observed for Qwen3-VL-8B in the main text.

\begin{table}[h]
  \centering
  \footnotesize
  \setlength{\tabcolsep}{4pt}
  \renewcommand{\arraystretch}{1.1}
  \caption{\textbf{InternVL3.5-8B short-form QA} (F1).}
  \label{tab:cross_arch_internvl_short}
  \begin{tabular}{@{}l cccc@{}}
    \toprule
    \textbf{Method} & \textbf{NQ} & \textbf{HotpotQA} & \textbf{NewsQA} & \textbf{TQA} \\
    \midrule
    Plain VQA           & 59.1                  & 31.5                  & 44.4                  & 66.5                 \\
    Random              & 59.3\up{0.3}          & 31.6\up{0.3}          & 44.6\up{0.5}          & 66.4\dn{0.2}         \\
    Uniform             & 59.6\up{0.8}          & 31.9\up{1.3}          & 44.9\up{1.1}          & 67.1\up{0.9}         \\
    VEA-Br              & 59.0\dn{0.2}          & 29.4\dn{6.7}          & 40.8\dn{8.1}          & 62.8\dn{5.6}         \\
    VEA-Con             & 59.9\up{1.4}          & 29.6\dn{6.0}          & 40.3\dn{9.2}          & 65.3\dn{1.8}         \\
    CGR                 & 33.6\dn{43.1}         & 19.8\dn{37.2}         & 29.0\dn{34.7}         & 35.0\dn{47.4}        \\
    VAR                 & 52.4\dn{11.3}         & 25.9\dn{17.7}         & 42.6\dn{4.1}          & 59.5\dn{10.5}        \\
    AGLA                & 53.0\dn{10.3}         & 30.2\dn{4.1}          & 40.6\dn{8.6}          & 61.7\dn{7.2}         \\
    \midrule
    \textbf{AGAR (ours)}& \textbf{67.4}\up{14.0}& \textbf{32.8}\up{4.1} & \textbf{47.2}\up{6.3} & \textbf{68.4}\up{2.9} \\
    \bottomrule
  \end{tabular}
\end{table}

\begin{table}[h]
  \centering
  \footnotesize
  \setlength{\tabcolsep}{3pt}
  \renewcommand{\arraystretch}{1.1}
  \caption{\textbf{InternVL3.5-8B multi-page benchmarks} (F1).}
  \label{tab:cross_arch_internvl_multi}
  \resizebox{\linewidth}{!}{%
  \begin{tabular}{@{}l cccc | cccc@{}}
    \toprule
    & \multicolumn{4}{c|}{\textbf{LongBench (multi-page) -- F1}}
    & \multicolumn{4}{c}{\textbf{LoCoMo (multi-page memory) -- F1}} \\
    \cmidrule(lr){2-5}\cmidrule(lr){6-9}
    \textbf{Method} & \textbf{LB-HQA} & \textbf{LB-2WMQA} & \textbf{LB-MSQ} & \textbf{LB-QP}
                    & \textbf{SingleHop} & \textbf{MultiHop} & \textbf{Temporal} & \textbf{OpenDom.} \\
    \midrule
    Plain VQA           & 48.6                  & 43.1                  & 24.0                  & 37.2
                        & 35.9                  & 26.0                  & 28.1                  & 16.4                 \\
    Random              & 47.5\dn{2.3}          & 42.9\dn{0.5}          & 23.8\dn{0.8}          & 37.3\up{0.3}
                        & 38.6\up{7.5}          & 26.9\up{3.5}          & 28.3\up{0.7}          & 17.2\up{4.9}         \\
    Uniform             & 48.6\eq               & 43.1\eq               & 24.0\eq               & 37.2\eq
                        & 35.8\dn{0.3}          & 26.0\eq               & 28.1\eq               & 16.4\eq              \\
    VEA-Br              & 48.3\dn{0.6}          & 31.3\dn{27.4}         & 7.1\dn{70.4}          & 21.0\dn{43.5}
                        & 15.3\dn{57.4}         & 13.1\dn{49.6}         & 31.4\up{11.7}         & 8.1\dn{50.6}         \\
    VEA-Con             & 37.0\dn{23.9}         & 32.9\dn{23.7}         & 10.1\dn{57.9}         & 22.4\dn{39.8}
                        & 17.2\dn{52.1}         & 15.0\dn{42.3}         & 31.1\up{10.7}         & 8.1\dn{50.6}         \\
    CGR                 & 24.9\dn{48.8}         & 23.8\dn{44.8}         & 9.5\dn{60.3}          & 19.5\dn{47.7}
                        & 12.2\dn{66.2}         & 10.3\dn{60.4}         & 5.7\dn{79.6}          & 9.9\dn{39.4}         \\
    VAR                 & 30.0\dn{38.2}         & 30.2\dn{30.0}         & 13.4\dn{44.0}         & 12.4\dn{66.7}
                        & 11.6\dn{67.8}         & 13.5\dn{47.9}         & 15.6\dn{44.4}         & 11.8\dn{28.0}        \\
    AGLA                & 31.7\dn{34.8}         & 29.5\dn{31.5}         & 17.8\dn{25.7}         & 17.9\dn{51.9}
                        & 15.3\dn{57.5}         & 14.6\dn{44.0}         & 16.6\dn{41.1}         & 14.9\dn{9.5}         \\
    \midrule
    \textbf{AGAR (ours)}& \textbf{48.9}\up{0.6} & \textbf{43.8}\up{1.6} & \textbf{25.6}\up{6.7} & \textbf{38.0}\up{2.2}
                        & \textbf{42.2}\up{17.5}& \textbf{27.6}\up{6.2} & \textbf{31.4}\up{11.7}& \textbf{17.8}\up{8.5}\\
    \bottomrule
  \end{tabular}%
  }
\end{table}

\begin{table}[h]
  \centering
  \footnotesize
  \setlength{\tabcolsep}{4pt}
  \renewcommand{\arraystretch}{1.1}
  \caption{\textbf{GLM-4.1V-9B-Thinking short-form QA} (F1).}
  \label{tab:cross_arch_glm_short}
  \begin{tabular}{@{}l cccc@{}}
    \toprule
    \textbf{Method} & \textbf{NQ} & \textbf{HotpotQA} & \textbf{NewsQA} & \textbf{TQA} \\
    \midrule
    Plain VQA           & 54.8                  & 40.9                  & 58.1                  & 65.1                 \\
    Random              & 54.7\dn{0.2}          & 41.0\up{0.2}          & 57.9\dn{0.3}          & 65.3\up{0.3}         \\
    Uniform             & 55.4\up{1.1}          & 41.4\up{1.2}          & 58.4\up{0.5}          & 65.7\up{0.9}         \\
    VEA-Br              & 43.9\dn{19.9}         & 39.2\dn{4.2}          & 58.0\dn{0.2}          & 66.4\up{2.0}         \\
    VEA-Con             & 47.3\dn{13.7}         & 40.7\dn{0.5}          & 58.4\up{0.5}          & 65.2\up{0.2}         \\
    CGR                 & 47.4\dn{13.5}         & 36.9\dn{9.8}          & 49.7\dn{14.5}         & 67.4\up{3.5}         \\
    VAR                 & 58.6\up{7.0}          & 36.9\dn{9.8}          & 57.3\dn{1.4}          & 67.6\up{3.8}         \\
    AGLA                & 55.9\up{2.0}          & 43.2\up{5.6}          & 49.5\dn{14.8}         & 67.2\up{3.2}         \\
    \midrule
    \textbf{AGAR (ours)}& \textbf{61.8}\up{12.8}& \textbf{48.0}\up{17.4}& \textbf{58.5}\up{0.7} & \textbf{68.1}\up{4.6}\\
    \bottomrule
  \end{tabular}
\end{table}

\begin{table}[h]
  \centering
  \footnotesize
  \setlength{\tabcolsep}{3pt}
  \renewcommand{\arraystretch}{1.1}
  \caption{\textbf{GLM-4.1V-9B-Thinking multi-page benchmarks} (F1).}
  \label{tab:cross_arch_glm_multi}
  \resizebox{\linewidth}{!}{%
  \begin{tabular}{@{}l cccc | cccc@{}}
    \toprule
    & \multicolumn{4}{c|}{\textbf{LongBench (multi-page) -- F1}}
    & \multicolumn{4}{c}{\textbf{LoCoMo (multi-page memory) -- F1}} \\
    \cmidrule(lr){2-5}\cmidrule(lr){6-9}
    \textbf{Method} & \textbf{LB-HQA} & \textbf{LB-2WMQA} & \textbf{LB-MSQ} & \textbf{LB-QP}
                    & \textbf{SingleHop} & \textbf{MultiHop} & \textbf{Temporal} & \textbf{OpenDom.} \\
    \midrule
    Plain VQA           & 44.5                  & 50.1                  & 24.8                  & 24.8
                        & 29.2                  & 17.8                  & 11.9                  & 13.7                 \\
    Random              & 43.5\dn{2.2}          & 46.5\dn{7.2}          & 25.5\up{2.8}          & 23.9\dn{3.6}
                        & 28.6\dn{2.1}          & 20.2\up{13.5}         & 11.5\dn{3.4}          & 9.4\dn{31.4}         \\
    Uniform             & 44.5\eq               & 50.2\up{0.2}          & 24.8\eq               & 24.2\dn{2.4}
                        & 29.2\eq               & 17.8\eq               & 11.9\eq               & 13.7\eq              \\
    VEA-Br              & 46.7\up{4.9}          & 50.2\up{0.2}          & 24.7\dn{0.4}          & 23.6\dn{4.8}
                        & 31.0\up{6.2}          & 17.8\eq               & 12.6\up{5.9}          & 10.5\dn{23.4}        \\
    VEA-Con             & 45.9\up{3.1}          & 49.9\dn{0.4}          & 23.9\dn{3.6}          & 24.9\up{0.4}
                        & 29.8\up{2.1}          & 18.3\up{2.8}          & 12.1\up{1.7}          & 13.3\dn{2.9}         \\
    CGR                 & 45.8\up{2.9}          & 24.7\dn{50.8}         & 21.9\dn{11.7}         & 19.9\dn{19.7}
                        & 27.5\dn{6.0}          & 14.8\dn{16.8}         & 4.9\dn{59.1}          & 7.2\dn{47.2}         \\
    VAR                 & 45.6\up{2.6}          & 50.3\up{0.4}          & 24.5\dn{1.2}          & 22.7\dn{8.4}
                        & 32.5\up{11.2}         & 19.4\up{8.9}          & 11.8\dn{0.7}          & 11.7\dn{14.4}        \\
    AGLA                & 45.7\up{2.7}          & 38.9\dn{22.4}         & 26.5\up{6.8}          & 18.6\dn{25.0}
                        & 32.2\up{10.3}         & 17.4\dn{2.2}          & 9.0\dn{24.4}          & 7.7\dn{43.8}         \\
    \midrule
    \textbf{AGAR (ours)}& \textbf{46.9}\up{5.4} & \textbf{50.7}\up{1.2} & \textbf{26.8}\up{8.1} & \textbf{25.9}\up{4.4}
                        & \textbf{32.7}\up{12.0}& \textbf{22.0}\up{23.6}& \textbf{13.5}\up{13.4}& \textbf{16.2}\up{18.2}\\
    \bottomrule
  \end{tabular}%
  }
\end{table}

\begin{table}[h]
  \centering
  \footnotesize
  \setlength{\tabcolsep}{4pt}
  \renewcommand{\arraystretch}{1.1}
  \caption{\textbf{Glyph short-form QA} (F1).}
  \label{tab:cross_arch_glyph_short}
  \begin{tabular}{@{}l cccc@{}}
    \toprule
    \textbf{Method} & \textbf{NQ} & \textbf{HotpotQA} & \textbf{NewsQA} & \textbf{TQA} \\
    \midrule
    Plain VQA           & 59.3                  & 56.4                  & 57.1                  & 75.8                 \\
    Random              & 63.0\up{6.2}          & 56.9\up{0.9}          & 55.3\dn{3.2}          & 74.4\dn{1.8}         \\
    Uniform             & 62.0\up{4.6}          & 53.4\dn{5.3}          & 56.3\dn{1.4}          & 73.4\dn{3.2}         \\
    VEA-Br              & 49.4\dn{16.7}         & 51.0\dn{9.6}          & 54.8\dn{4.0}          & 70.2\dn{7.4}         \\
    VEA-Con             & 50.5\dn{14.8}         & 50.5\dn{10.5}         & 54.5\dn{4.6}          & 71.0\dn{6.3}         \\
    CGR                 & 56.2\dn{5.2}          & 57.1\up{1.2}          & 47.0\dn{17.6}         & 76.1\up{0.4}         \\
    VAR                 & 66.9\up{12.8}         & 52.9\dn{6.2}          & 55.3\dn{3.2}          & 75.9\up{0.1}         \\
    AGLA                & 62.0\up{4.6}          & 49.2\dn{12.7}         & 49.6\dn{13.2}         & 75.7\dn{0.1}         \\
    \midrule
    \textbf{AGAR (ours)}& \textbf{67.8}\up{14.3}& \textbf{59.6}\up{5.7} & \textbf{60.1}\up{5.3} & \textbf{77.3}\up{2.0}\\
    \bottomrule
  \end{tabular}
\end{table}

\begin{table}[h]
  \centering
  \footnotesize
  \setlength{\tabcolsep}{3pt}
  \renewcommand{\arraystretch}{1.1}
  \caption{\textbf{Glyph multi-page benchmarks} (F1).}
  \label{tab:cross_arch_glyph_multi}
  \resizebox{\linewidth}{!}{%
  \begin{tabular}{@{}l cccc | cccc@{}}
    \toprule
    & \multicolumn{4}{c|}{\textbf{LongBench (multi-page) -- F1}}
    & \multicolumn{4}{c}{\textbf{LoCoMo (multi-page memory) -- F1}} \\
    \cmidrule(lr){2-5}\cmidrule(lr){6-9}
    \textbf{Method} & \textbf{LB-HQA} & \textbf{LB-2WMQA} & \textbf{LB-MSQ} & \textbf{LB-QP}
                    & \textbf{SingleHop} & \textbf{MultiHop} & \textbf{Temporal} & \textbf{OpenDom.} \\
    \midrule
    Plain VQA           & 61.9                  & 68.2                  & 45.7                  & 38.4
                        & 42.0                  & 29.1                  & 21.2                  & 23.0                 \\
    Random              & 61.7\dn{0.3}          & 70.3\up{3.1}          & 50.9\up{11.4}         & 37.2\dn{3.1}
                        & 43.0\up{2.4}          & 26.9\dn{7.6}          & 19.0\dn{10.4}         & 20.9\dn{9.1}         \\
    Uniform             & 62.0\up{0.2}          & 66.1\dn{3.1}          & 45.7\eq               & 36.2\dn{5.7}
                        & 42.0\eq               & 29.1\eq               & 21.2\eq               & 23.1\up{0.4}         \\
    VEA-Br              & 11.6\dn{81.3}         & 65.9\dn{3.4}          & 41.6\dn{9.0}          & 37.4\dn{2.6}
                        & 41.4\dn{1.4}          & 27.2\dn{6.5}          & 19.8\dn{6.6}          & 18.5\dn{19.6}        \\
    VEA-Con             & 60.4\dn{2.4}          & 66.3\dn{2.8}          & 43.0\dn{5.9}          & 38.0\dn{1.0}
                        & 43.0\up{2.4}          & 30.4\up{4.5}          & 22.2\up{4.7}          & 18.2\dn{20.9}        \\
    CGR                 & 61.0\dn{1.5}          & 52.1\dn{23.7}         & 50.8\up{11.2}         & 39.0\up{1.6}
                        & 36.5\dn{13.1}         & 27.3\dn{6.2}          & 15.7\dn{26.0}         & 14.6\dn{36.6}        \\
    VAR                 & 60.4\dn{2.4}          & 69.1\up{1.3}          & 45.4\dn{0.6}          & 38.4\eq
                        & 43.2\up{2.9}          & 29.4\up{1.0}          & 22.0\up{3.8}          & 21.5\dn{6.4}         \\
    AGLA                & 56.6\dn{8.5}          & 50.0\dn{26.7}         & 44.0\dn{3.7}          & 31.4\dn{18.1}
                        & 38.5\dn{8.4}          & 28.0\dn{3.8}          & 21.9\up{3.3}          & 18.6\dn{19.0}        \\
    \midrule
    \textbf{AGAR (ours)}& \textbf{63.6}\up{2.7} & \textbf{72.5}\up{6.3} & \textbf{52.2}\up{14.2}& \textbf{40.4}\up{5.2}
                        & \textbf{44.5}\up{6.0} & \textbf{30.7}\up{5.5} & \textbf{23.2}\up{9.4} & \textbf{24.3}\up{5.7} \\
    \bottomrule
  \end{tabular}%
  }
\end{table}

\section{Layer-wise Evidence Attention on Additional Datasets}
\label{app:relative_attention_appendix}

\begin{figure}[h]
  \centering
  \includegraphics[width=\linewidth]{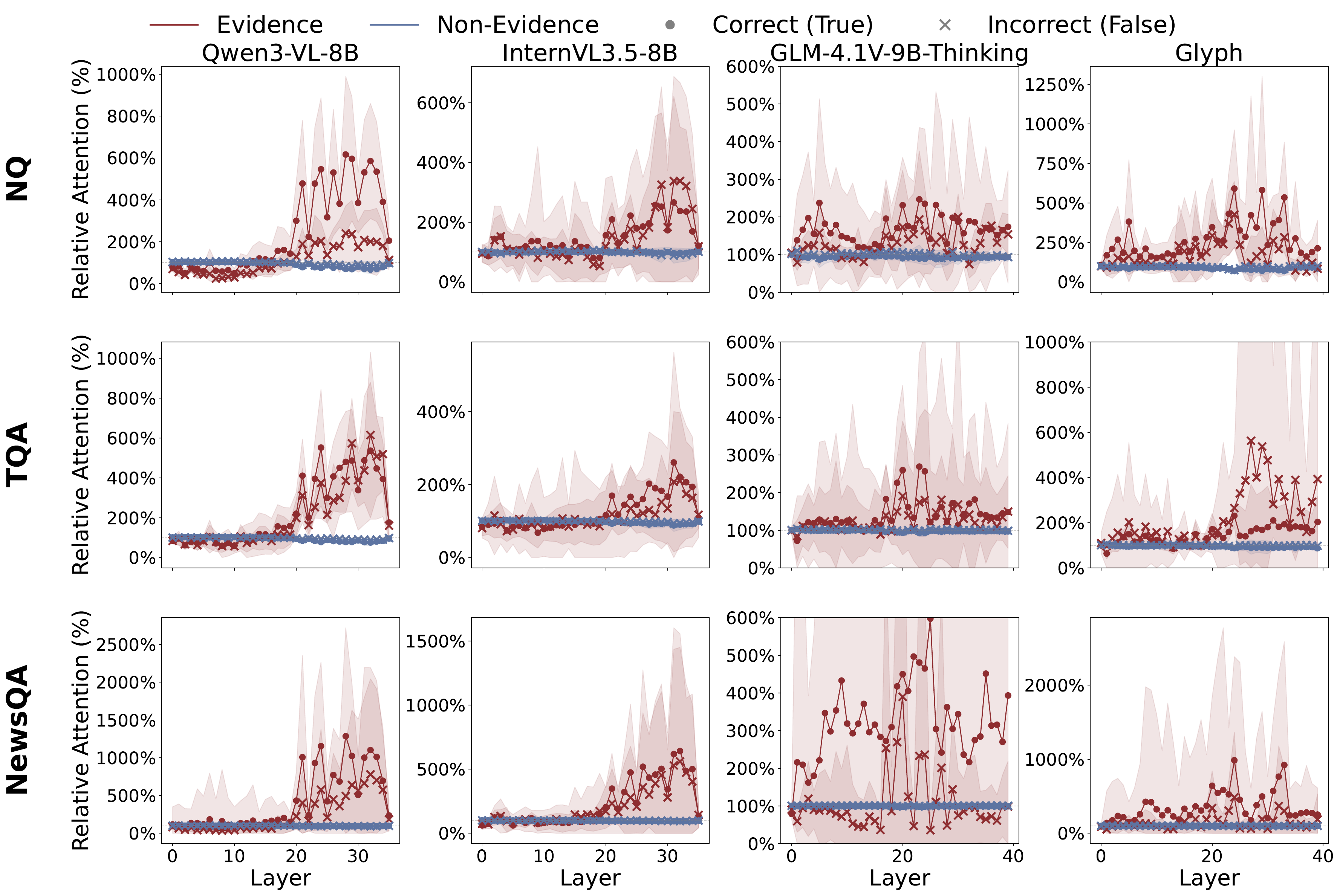}
  \caption{\textbf{Layer-wise relative attention on NQ / TriviaQA /
  NewsQA (appendix companion to Fig.~\ref{fig:relative_attention_panel})}.}
  \label{fig:relative_attention_panel_appendix}
\end{figure}

Fig.~\ref{fig:relative_attention_panel_appendix} extends the
HotpotQA panel of Fig.~\ref{fig:relative_attention_panel} to NQ /
TriviaQA / NewsQA across the same four VLMs (Qwen3-VL-8B,
InternVL3.5-8B, GLM-4.1V-9B-Thinking, and Glyph). Each cell shows
per-layer relative attention to Evidence vs.\ Non-Evidence image
tokens, with samples split by answer correctness (Correct / Incorrect)
and aggregated as mean$\pm$1\,std. The flat-then-rising profile
observed on HotpotQA --- Evidence attention is roughly equal to
Non-Evidence in shallow layers and pulls sharply ahead from the
late-middle layers onward --- replicates across all three additional
datasets and all four models.

\section{Per-head NDCG Heatmaps on Additional Datasets}
\label{app:per_head_ndcg_appendix}

\begin{figure}[h]
  \centering
  \includegraphics[width=\linewidth]{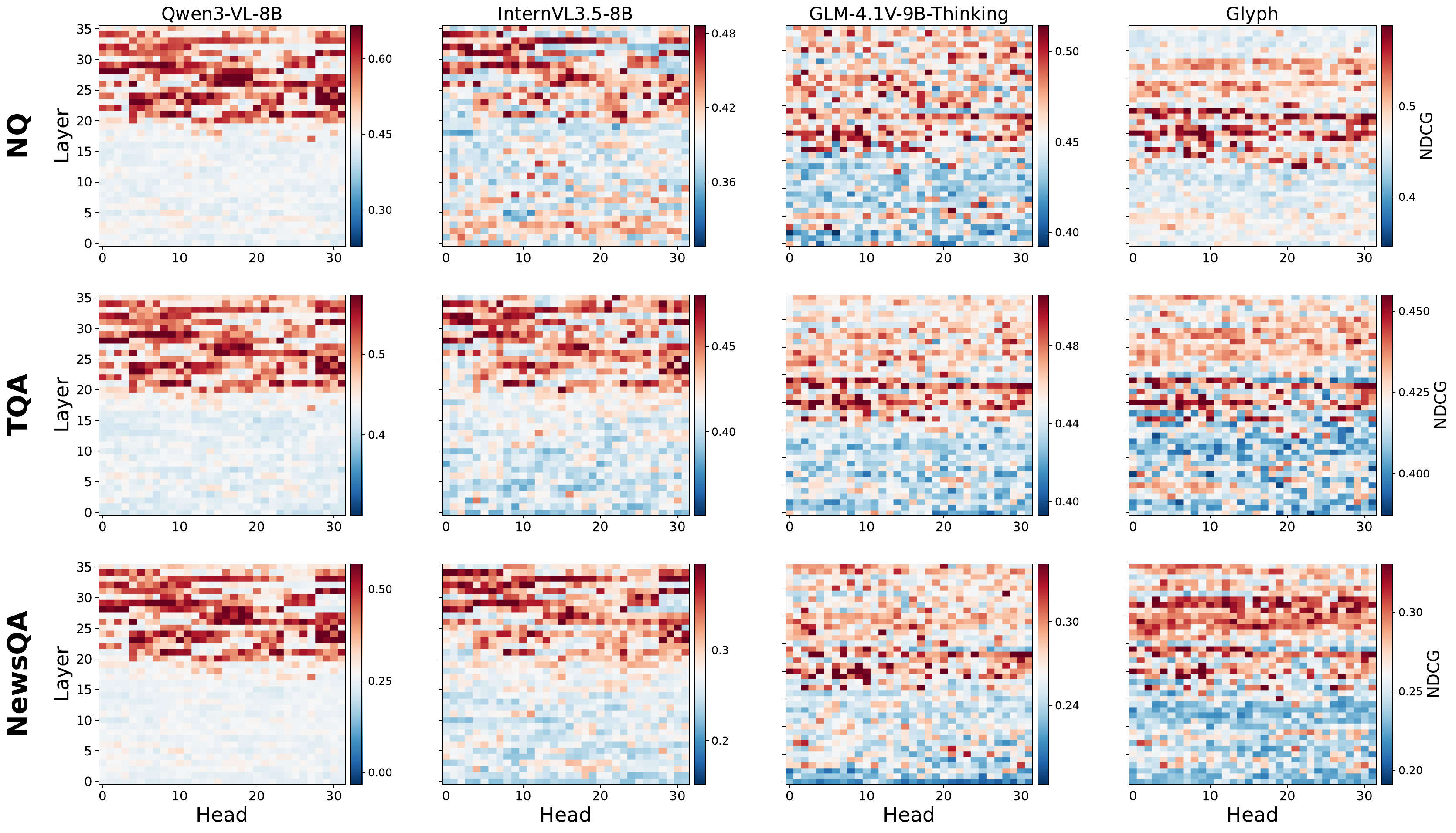}
  \caption{\textbf{Per-head NDCG of last-token attention to evidence
  tokens on NQ / TriviaQA / NewsQA (appendix companion to
  Fig.~\ref{fig:per_head_ndcg_heatmap})}.}
  \label{fig:per_head_ndcg_heatmap_appendix}
\end{figure}

Fig.~\ref{fig:per_head_ndcg_heatmap_appendix} extends the HotpotQA panel
of Fig.~\ref{fig:per_head_ndcg_heatmap} to NQ / TriviaQA / NewsQA. Each
cell shows the per-head NDCG of last-token attention to evidence tokens
(layer $\times$ head), with the colour scale centred on the per-cell
median NDCG so that white maps to median and red/blue mark heads above
or below it. The same population of late-middle--layer evidence-attending
heads observed on HotpotQA is visible across all three additional
datasets and all four VLMs.

\begin{figure}[h]
  \centering
  \includegraphics[width=\linewidth]{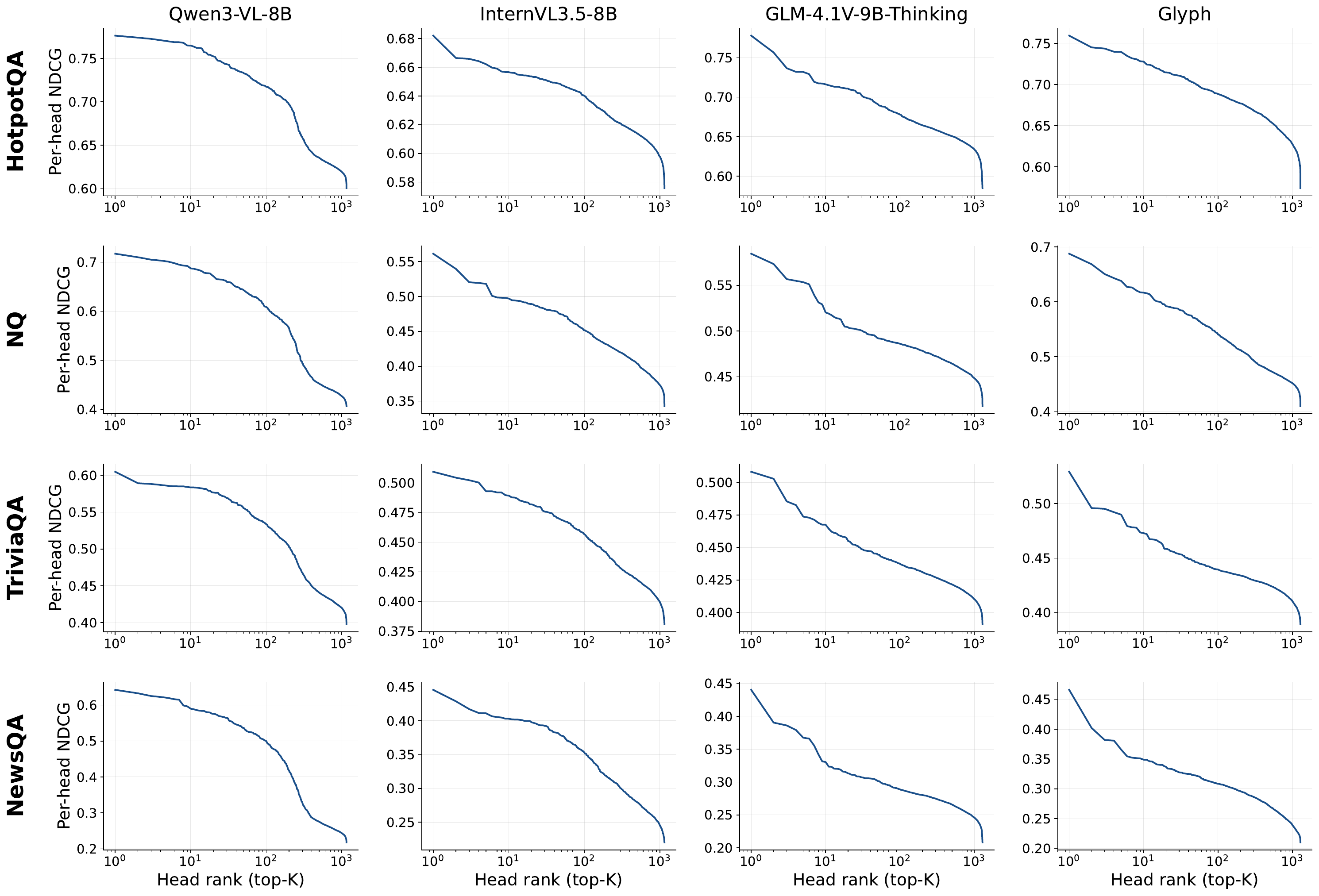}
  \caption{\textbf{Per-head NDCG vs.\ head rank} for the same four VLMs
  $\times$ four datasets as Fig.~\ref{fig:per_head_ndcg_heatmap_appendix}.
  Each curve sorts the (layer, head) heads by mean NDCG (descending) on
  a log-scale rank axis; the steep drop near the top and slow decay
  toward the median quantify the head-population structure visible in
  the layer\,$\times$\,head heatmaps.}
  \label{fig:per_head_rank_curve_panel}
\end{figure}

\section{Layer-range Aggregated NDCG}
\label{app:layer_range_ndcg}

Table~\ref{tab:layer_range_ndcg} reports the NDCG of last-token attention
to evidence patches after pooling head-averaged attention over different
ranges of decoder layers: the first half of layers (\emph{Early},
$0$--$50\%$ depth), the second half (\emph{Late}, $50$--$100\%$), and
the full stack (\emph{All}, $0$--$100\%$). Across every (model, dataset) cell, the late-half pooling
yields the highest NDCG, confirming that evidence-localizing attention
is concentrated in the second half of the network and that aggregating
the early half dilutes rather than reinforces the signal.

\begin{table}[h]
  \centering
  \footnotesize
  \setlength{\tabcolsep}{6pt}
  \renewcommand{\arraystretch}{1.1}
  \caption{\textbf{Layer-range aggregated NDCG} (mean over $n{=}100$
  samples per cell). Each cell is the NDCG of last-token attention,
  head-averaged and then pooled over the specified depth range. The
  Late ($50$--$100\%$) pool dominates in every (model, dataset) cell.}
  \label{tab:layer_range_ndcg}
  \begin{tabular}{@{}llccc@{}}
    \toprule
    \textbf{Model} & \textbf{Dataset} & \textbf{Early ($0$--$50\%$)}
      & \textbf{Late ($50$--$100\%$)} & \textbf{All ($0$--$100\%$)} \\
    \midrule
    \multirow{4}{*}{Qwen3-VL-8B}
        & NQ        & 0.439 & \textbf{0.606} & 0.535 \\
        & TriviaQA  & 0.428 & \textbf{0.540} & 0.507 \\
        & NewsQA    & 0.264 & \textbf{0.469} & 0.396 \\
        & HotpotQA  & 0.634 & \textbf{0.709} & 0.689 \\
    \midrule
    \multirow{4}{*}{InternVL3.5-8B}
        & NQ        & 0.392 & \textbf{0.454} & 0.435 \\
        & TriviaQA  & 0.402 & \textbf{0.466} & 0.440 \\
        & NewsQA    & 0.264 & \textbf{0.412} & 0.359 \\
        & HotpotQA  & 0.609 & \textbf{0.643} & 0.629 \\
    \midrule
    \multirow{4}{*}{GLM-4.1V-9B-Thinking}
        & NQ        & 0.455 & \textbf{0.468} & 0.460 \\
        & TriviaQA  & 0.417 & \textbf{0.434} & 0.424 \\
        & NewsQA    & 0.263 & \textbf{0.290} & 0.274 \\
        & HotpotQA  & 0.610 & \textbf{0.636} & 0.620 \\
    \midrule
    \multirow{4}{*}{Glyph}
        & NQ        & 0.497 & \textbf{0.555} & 0.520 \\
        & TriviaQA  & 0.415 & \textbf{0.441} & 0.422 \\
        & NewsQA    & 0.260 & \textbf{0.317} & 0.277 \\
        & HotpotQA  & 0.612 & \textbf{0.645} & 0.621 \\
    \bottomrule
  \end{tabular}
\end{table}

\section{Baseline Implementation}
\label{app:baseline_impl}

\textbf{\CGR} operates in two stages. First, the model is prompted with the rendered image and question to extract relevant content as structured textual evidence (verbatim snippets, names, dates, numbers, and brief surrounding context). Second, this extracted text is concatenated with the original question and passed to a text-only stage that produces the final answer, abstaining when evidence is insufficient.

\textbf{\VAR} constructs a binary visual mask from the model's final-layer attention scores. We aggregate attention across heads, select the globally highest-scoring patches, and present the model with a version of the image in which those patches are highlighted and the remainder is unchanged.

\textbf{\AGLA} is run using the official codebase and default hyperparameters. The method produces a question-conditioned saliency map, masks out low-relevance regions to form a local-evidence image, and ensembles generations from the original and masked images via AGLA decoding. No AGLA-specific parameters were tuned on our evaluation sets.

\textbf{VEA-Br / VEA-Con} are the brightness and contrast variants of Visual Evidence Augmentation~\citep{liu2025seeing}. We follow the default hyperparameters described in the paper (Appendix A.3, ``VEA and Baselines Implementation Details''): attention is aggregated by layer-span mean over the middle-to-late half ($[0.5, 1.0]$), with $\sigma{=}0.1$, $\alpha{=}0.9$, offset $=1.0$. VEA-Br up-weights pixel brightness in attended regions while VEA-Con increases local contrast; in both cases the augmented image replaces the plain page for a single forward pass.

\section{Impact on Base Token Font}
\label{app:token_base}
\label{app:qwen-tokenbase-figures}

\providecommand{\PlainVQA}{\textsc{PlainVQA}}
\providecommand{\FixedTB}{\textsc{Fixed}}
\providecommand{\RelativeTB}{\textsc{Relative}}
\providecommand{\TokenBase}{\textsc{TokenBase}}

\begin{figure}[t]
\centering
\includegraphics[width=\linewidth]{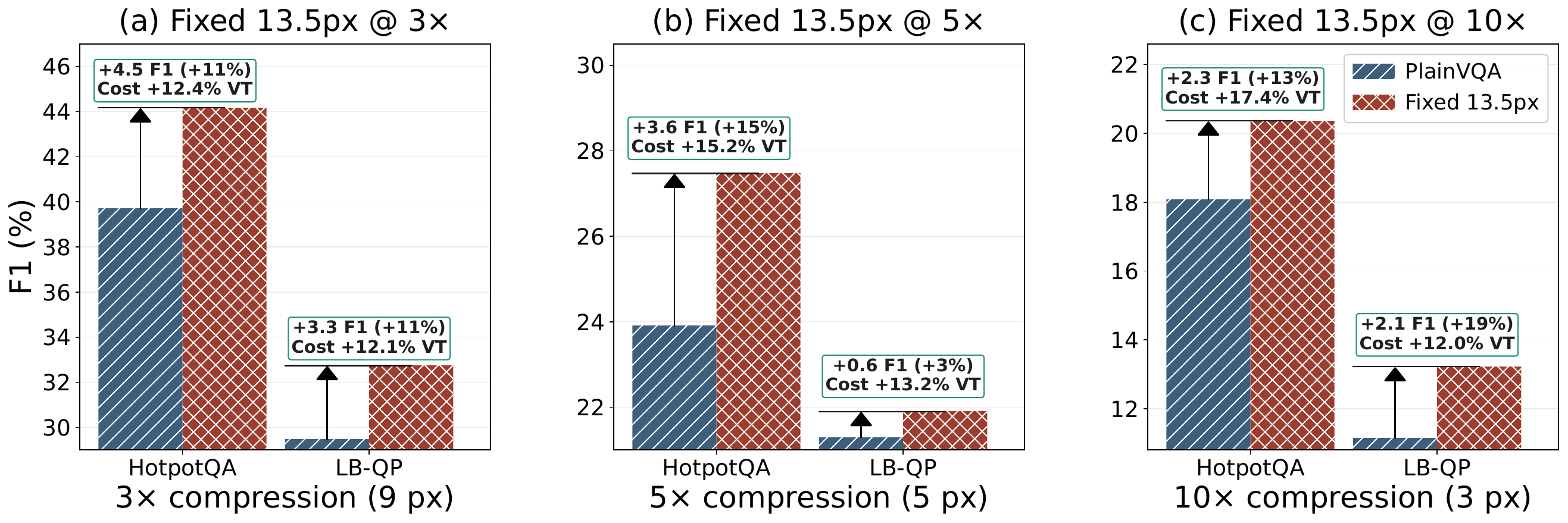}
\caption{\textbf{Fixed readable-size enlargement on Qwen3-VL-8B under
vision-token compression.} Blue: \PlainVQA{} F1; red: \FixedTB{} F1.
Panel (a) compares the Qwen \PlainVQA{} baseline at 3$\times$ compression
(9\,px) with \FixedTB{} under the same rendering, where fixed 13.5\,px equals
scale $1.5{\times}$; panels (b) and (c) show \FixedTB{} at 5$\times$
compression (5\,px) and 10$\times$ compression (3\,px).  Boxes report absolute
and relative $\Delta$F1 plus vision-token cost.}
\label{fig:qwen-tokenbase-fixed}
\end{figure}

\begin{figure}[t]
\centering
\includegraphics[width=\linewidth]{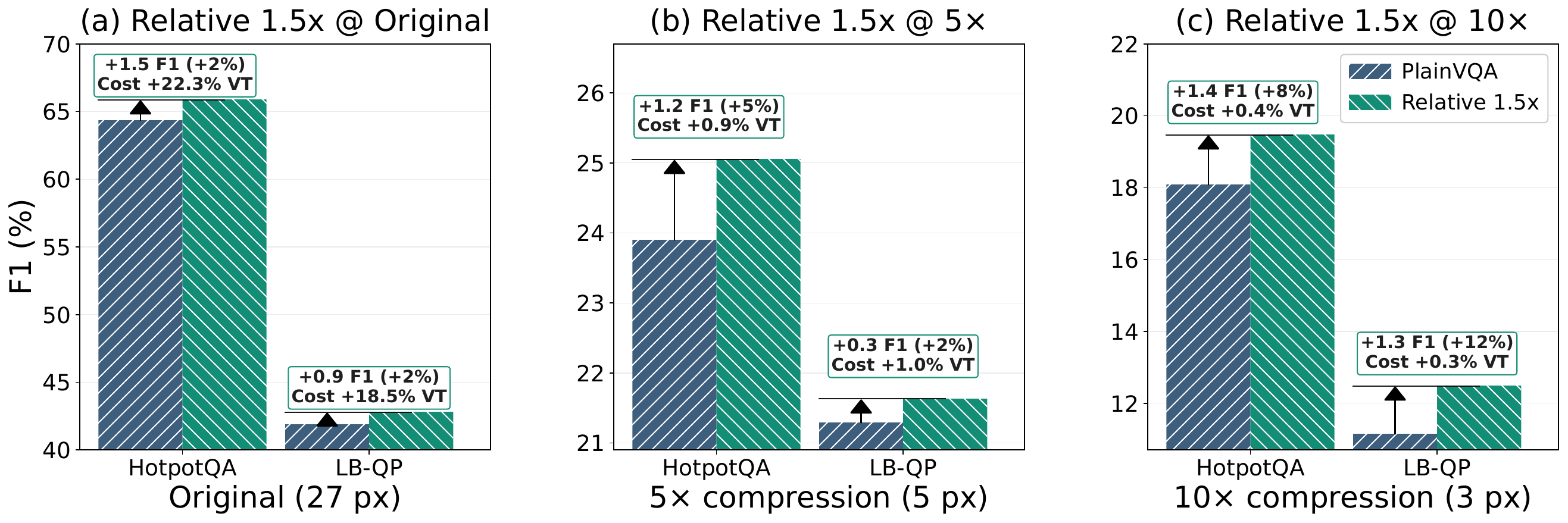}
\caption{\textbf{Relative-scale enlargement on Qwen3-VL-8B under vision-token
compression.} Blue: \PlainVQA{} F1; green: \RelativeTB{} F1.  Panel (a) shows
\RelativeTB{} on the original 27\,px rendering; panels (b) and (c) show
\RelativeTB{} at 5$\times$ compression (5\,px) and 10$\times$ compression
(3\,px).  Boxes report absolute and relative $\Delta$F1 plus vision-token
cost.}
\label{fig:qwen-tokenbase-relative}
\end{figure}

This section studies the impact of the base token font on AGAR. We evaluate
Qwen3-VL-8B-Instruct on HotpotQA and LB-QP under two degraded renderings:
5$\times$ compression (5\,px base font size) and 10$\times$ compression
(3\,px base font size). Figures~\ref{fig:qwen-tokenbase-fixed}
and~\ref{fig:qwen-tokenbase-relative} report the results under two enlargement
rules. Both variants select the top 2\% of attention patches and differ only in
the enlargement rule:

\begin{itemize}
    \item \FixedTB{} uses a fixed target: selected regions are redrawn at
    13.5\,px no matter how small the compressed base font size is.  This spends
    more visual tokens but explicitly targets a readable font size.
    \item \RelativeTB{} uses a relative target: selected regions are enlarged by
    $1.5{\times}$ from their current base font size, yielding 40.5\,px for the
    original 27\,px rendering, 7.5\,px under 5$\times$ compression, and 4.5\,px
    under 10$\times$ compression.  This keeps the compressed-setting token
    overhead very small.
\end{itemize}

The fixed-size figure also includes a 3$\times$ compression panel (9\,px base
font size; HotpotQA $n{=}7404$, LB-QP $n{=}200$), where the fixed 13.5\,px target
is exactly a $1.5{\times}$ enlargement of the selected 9\,px regions.  The
relative-size figure includes the original 27\,px rendering as a readable
reference point for the same relative rule.

\paragraph{Takeaway.}
Across all compression ratios and both enlargement rules, AGAR consistently
improves F1 over compressed \PlainVQA{}.

\section{Case Study}
\label{app:case_study}

\begin{figure}[h]
  \centering
  \includegraphics[width=\linewidth]{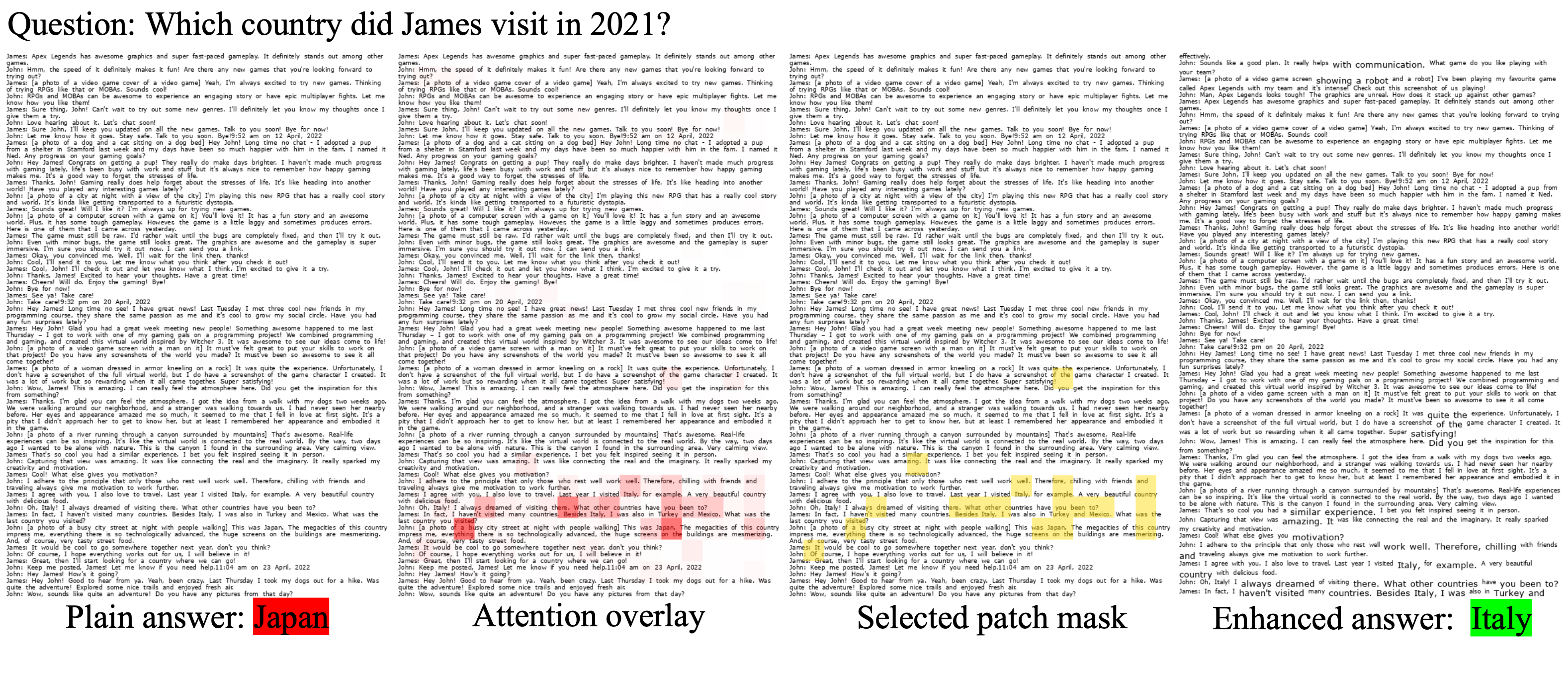}
  \caption{LoCoMo case for James's 2021 trip. \textbf{Evidence:} in a March 2022 dialogue, James says he visited Italy last year, while Japan is John's later trip. The enhanced answer changes Japan to Italy.}
  \label{fig:cs_first}
\end{figure}

\begin{figure}[h]
  \centering
  \includegraphics[width=\linewidth]{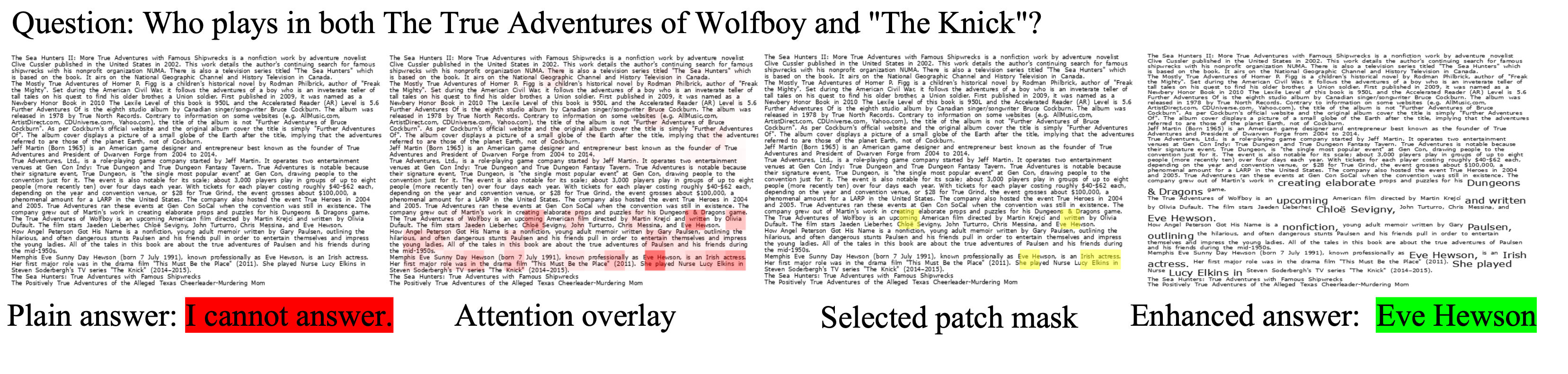}
  \caption{HotpotQA case on the actor shared by \emph{The True Adventures of Wolfboy} and \emph{The Knick}. \textbf{Evidence:} \emph{Wolfboy} stars Eve Hewson, and Eve Hewson played Nurse Lucy Elkins in \emph{The Knick}. The enhanced answer recovers Eve Hewson from the plain no-answer response.}
  \label{fig:cs_last}
\end{figure}

We provide case studies on various datasets from Figure~\ref{fig:cs_first} to Figure~\ref{fig:cs_last}.

\begin{figure}[h]
  \centering
  \includegraphics[width=\linewidth]{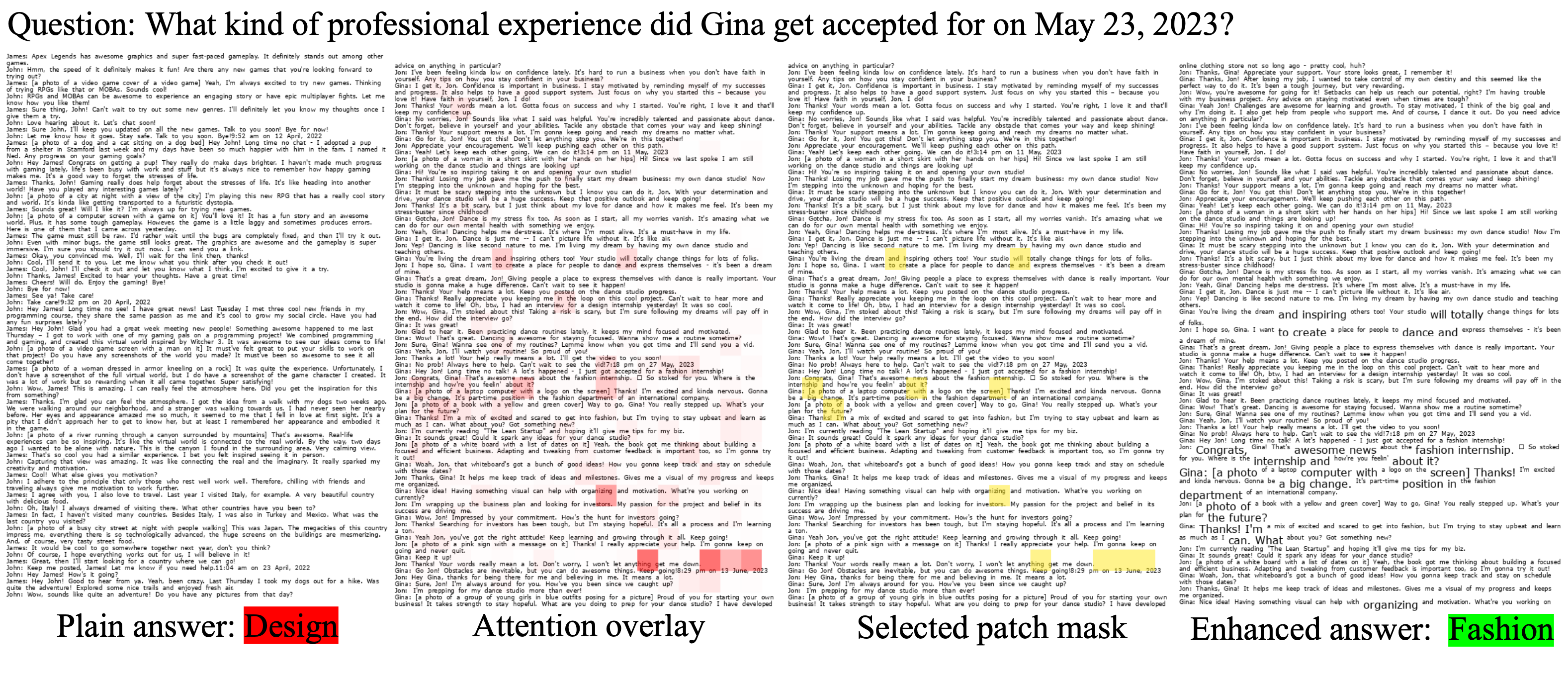}
  \caption{LoCoMo case for Gina's accepted professional experience. \textbf{Evidence:} Gina says she got accepted for a fashion internship; the design internship is only an earlier interview. The enhanced answer changes design internship to fashion internship.}
  \label{fig:}
\end{figure}

\begin{figure}[h]
  \centering
  \includegraphics[width=\linewidth]{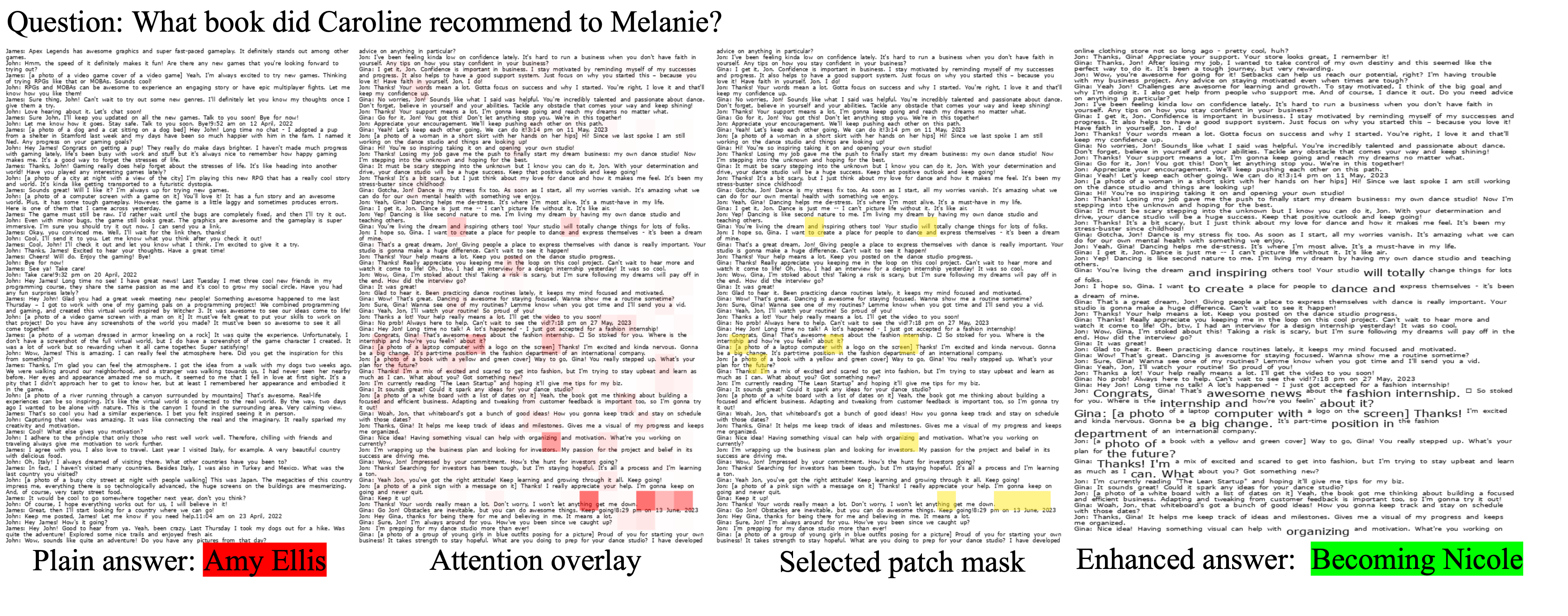}
  \caption{LoCoMo case for Caroline's book recommendation. \textbf{Evidence:} Caroline says she loved \emph{Becoming Nicole} by Amy Ellis Nutt and highly recommends it. The enhanced answer corrects the title to \emph{Becoming Nicole}.}
\end{figure}

\begin{figure}[h]
  \centering
  \includegraphics[width=\linewidth]{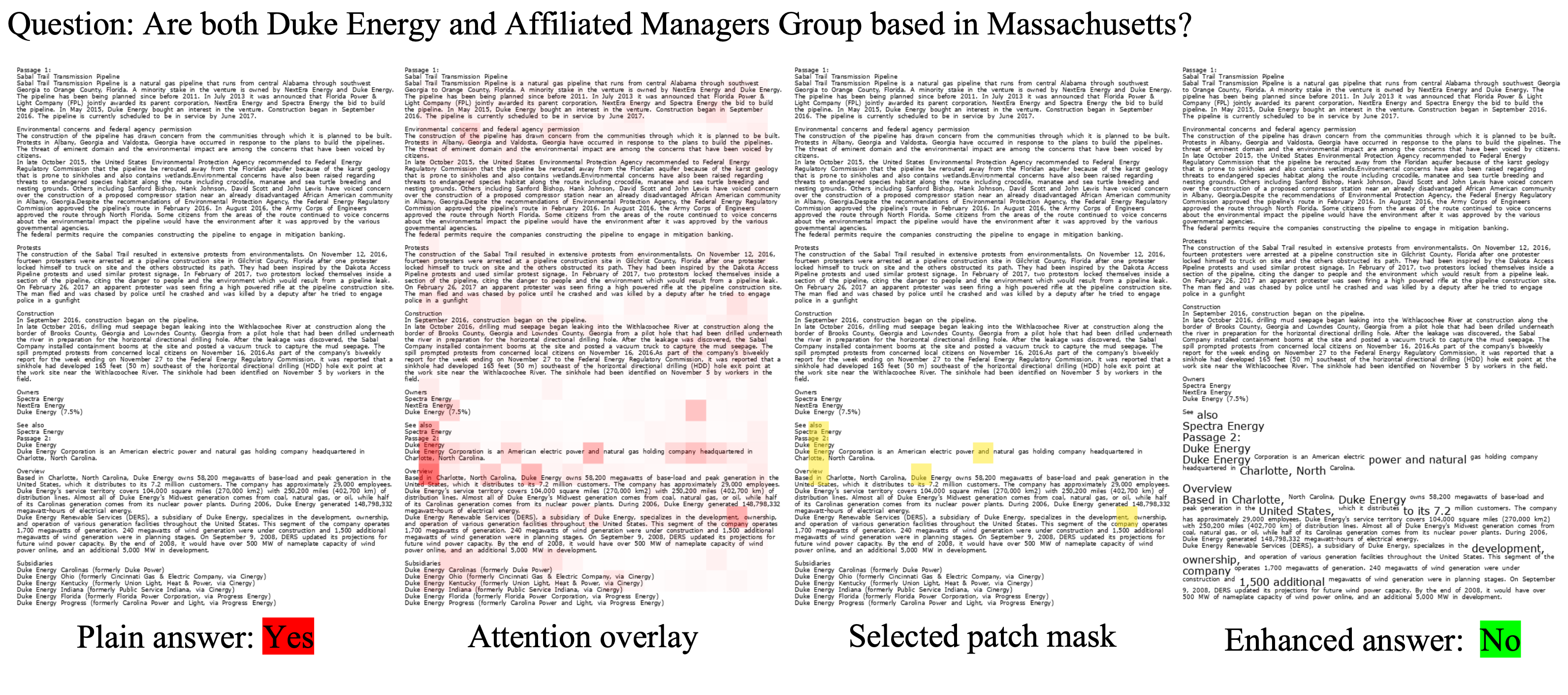}
  \caption{LongBench HotpotQA case asking whether Duke Energy and Affiliated Managers Group are both based in Massachusetts. \textbf{Evidence:} Duke Energy is headquartered in Charlotte, North Carolina, so the conjunction is false even though Affiliated Managers Group has a Massachusetts office. The enhanced answer changes Yes to No.}
  \label{fig:}
\end{figure}

\begin{figure}[h]
  \centering
  \includegraphics[width=\linewidth]{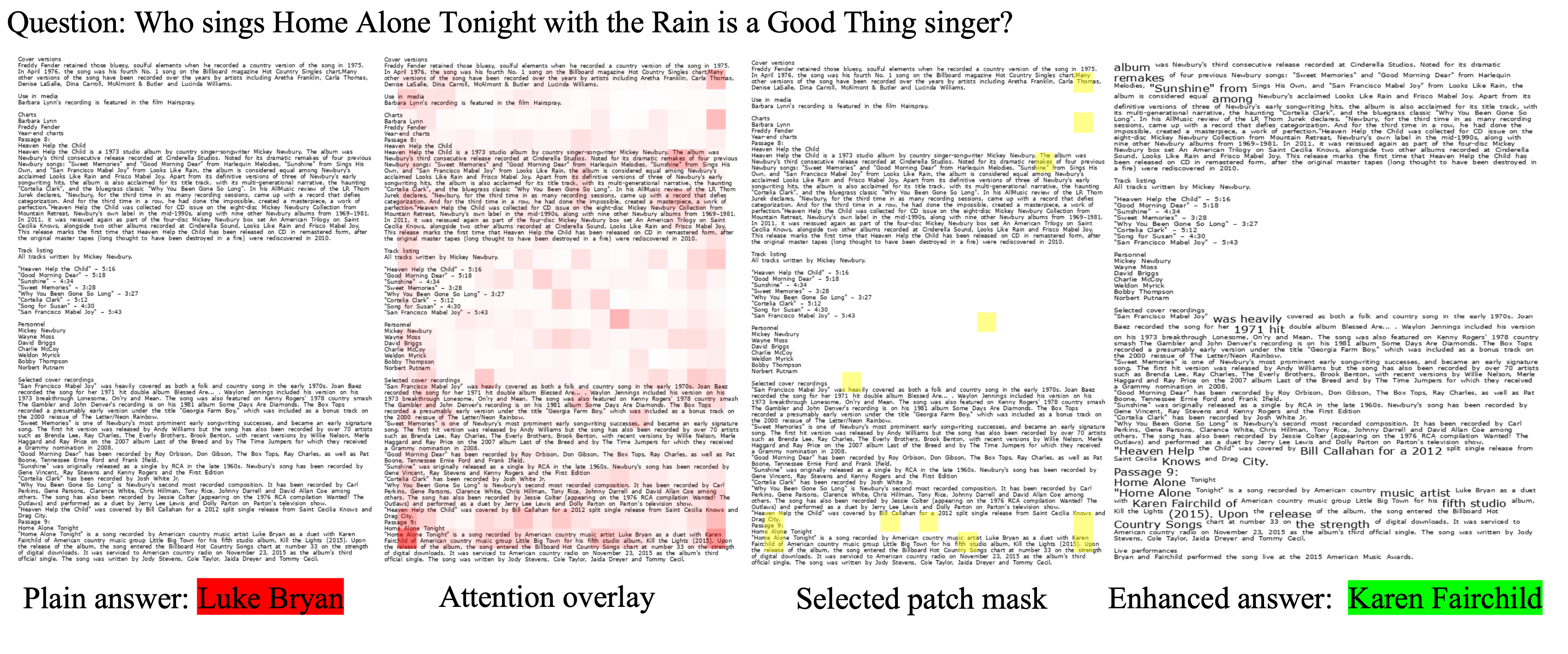}
  \caption{LongBench MuSiQue case linking \emph{Rain Is a Good Thing} to \emph{Home Alone Tonight}. \textbf{Evidence:} \emph{Rain Is a Good Thing} is by Luke Bryan, and \emph{Home Alone Tonight} is Luke Bryan's duet with Karen Fairchild. The enhanced answer changes Luke Bryan to Karen Fairchild.}
  \label{fig:}
\end{figure}

\begin{figure}[h]
  \centering
  \includegraphics[width=\linewidth]{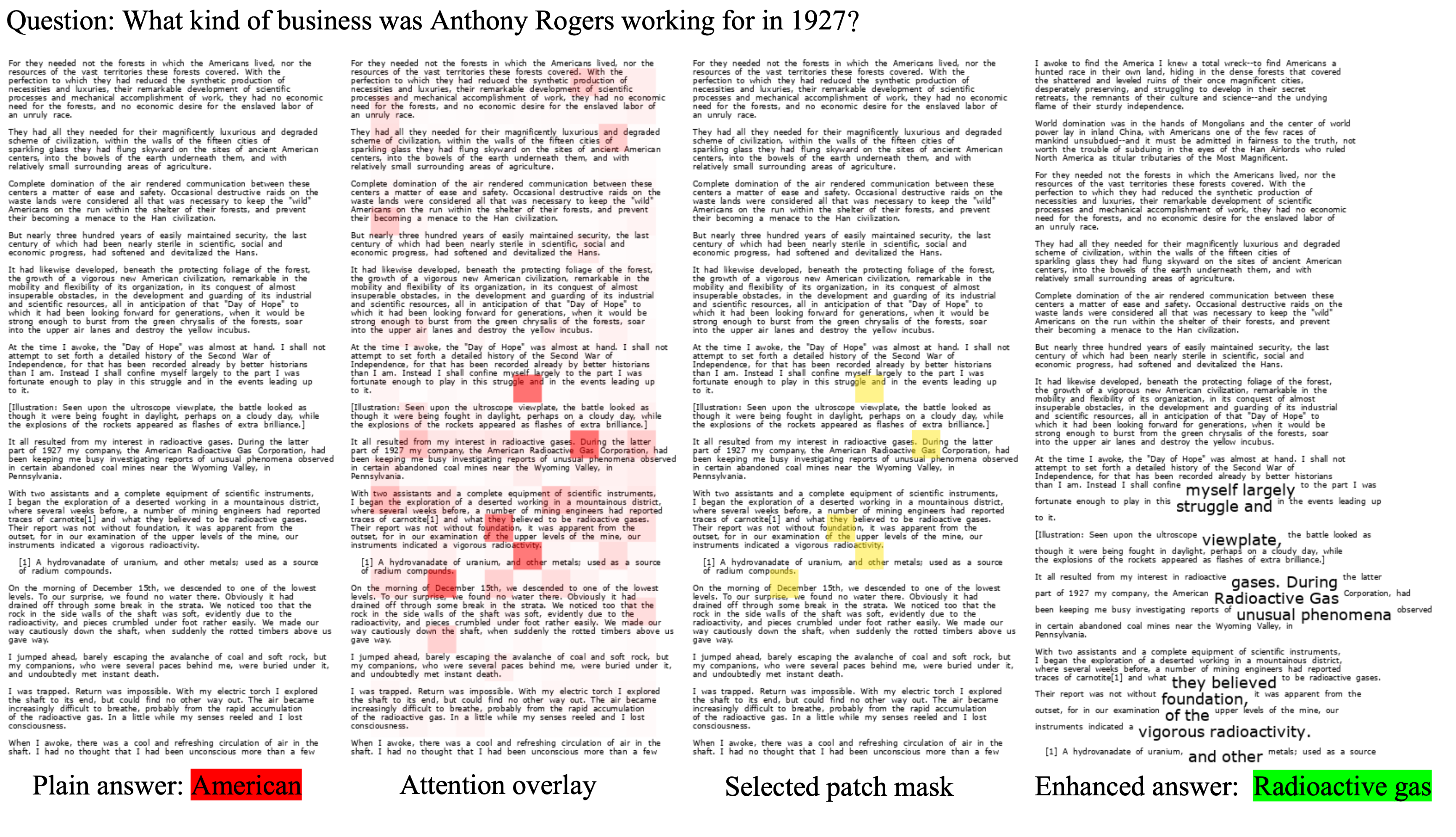}
  \caption{LongBench NarrativeQA case on Anthony Rogers's 1927 business. \textbf{Evidence:} the story says his company was the American Corporation. The enhanced answer gives the requested business type, radioactive gas, instead of the broader name.}
  \label{fig:}
\end{figure}

\begin{figure}[h]
  \centering
  \includegraphics[width=\linewidth]{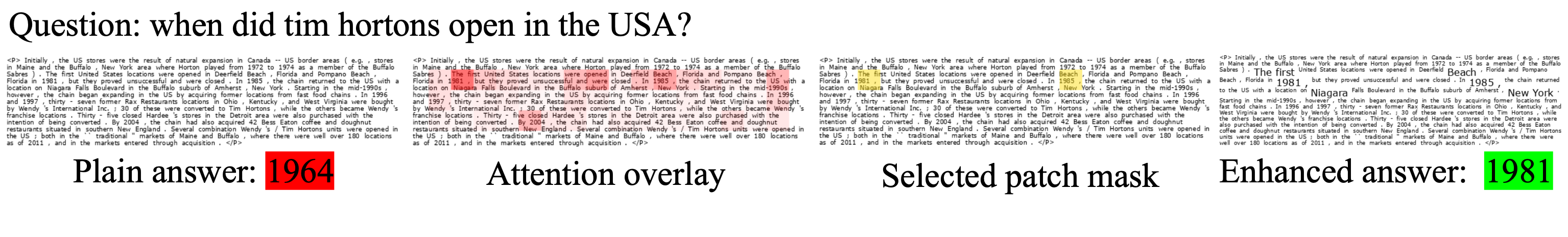}
  \caption{Natural Questions case on when Tim Hortons opened in the USA. \textbf{Evidence:} the first United States locations opened in Deerfield Beach and Pompano Beach, Florida, in 1981. The enhanced answer changes 1964 to the U.S. opening year, 1981.}
  \label{fig:}
\end{figure}

\begin{figure}[h]
  \centering
  \includegraphics[width=\linewidth]{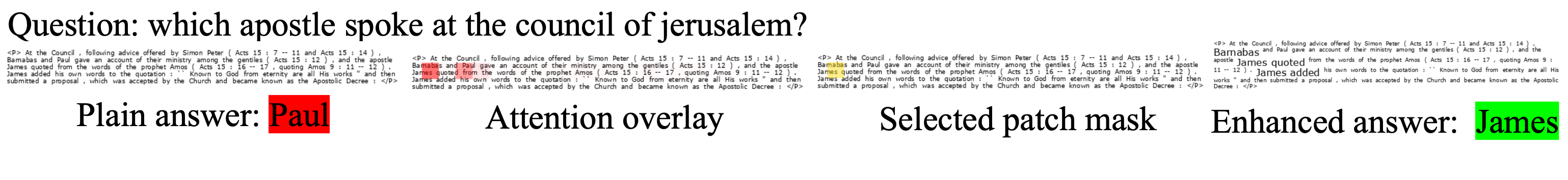}
  \caption{Natural Questions case on which apostle spoke at the Council. \textbf{Evidence:} Barnabas and Paul report their ministry, but the apostle James quotes Amos and submits the accepted proposal. The enhanced answer changes Paul to James.}
  \label{fig:}
\end{figure}

\begin{figure}[h]
  \centering
  \includegraphics[width=\linewidth]{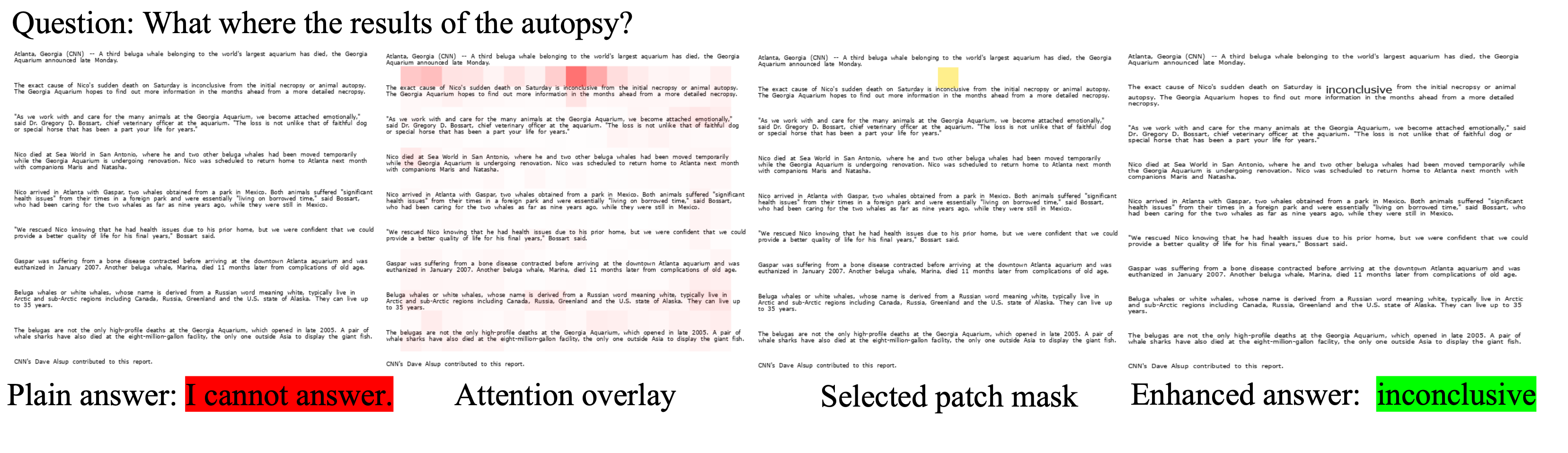}
  \caption{NewsQA case on the autopsy result. \textbf{Evidence:} the initial necropsy, or animal autopsy, found Nico's cause of death inconclusive. The enhanced answer changes the plain no-answer response to inconclusive.}
  \label{fig:}
\end{figure}

\section{Experiments Compute Resources}
\label{app:compute}

All experiments are run on Python 3.10 with CUDA 12.8 and PyTorch in
\texttt{bfloat16}. AGAR is inference-only and does not require training,
so the dominant cost is two forward passes per question (one to read
attention, one on the re-rendered page). A single GPU with at least
50\,GB of free memory is recommended (e.g.\ NVIDIA A100-80G, H100, or
H200); on shorter inputs a 40\,GB card is sufficient, while the
multi-page LongBench / LoCoMo settings benefit from the larger memory
to fit the full visual-token sequence. Our experiments were carried out
on a 2$\times$ A100-80G server.

\section{Broader Impact}
\label{app:broader_impact}

AGAR is a training-free, plug-and-play enhancement on top of an existing
vision--language model: it does not collect new data, does not fine-tune
weights, and does not change the prompt. Its effect is to redistribute
visual capacity on a re-rendered page so that the model's own attention
decides which words appear at a more legible scale before the second
forward pass. We discuss potential broader impacts at a high level
below, and we keep the discussion deliberately broad rather than
tied to specific deployments.

\paragraph{Potential positive impacts.}
By making vision-text comprehension more reliable on long, dense, or
visually compressed pages, AGAR can plausibly contribute to more
useful document, screenshot, and memory QA assistants under tight
context-window or token budgets, including settings where readers
benefit from larger, more salient text (e.g.\ accessibility-oriented
or low-bandwidth scenarios). Because the method exposes which patches
the model itself attended to, the resulting magnified renderings also
make the model's internal evidence selection more inspectable, which
may support debugging, auditing, and downstream
interpretability research on how VLMs read visualized text.

\paragraph{Potential negative impacts.}
AGAR inherits all general risks of VLM-based question answering:
hallucinated answers, biased or unsafe outputs, leakage of sensitive
content present in the input page, and uneven reliability across
languages, domains, scripts, and user populations. Because the method
is built on the model's own attention, any systematic bias in what
the underlying VLM tends to attend to will likewise be magnified
on the re-rendered page, which could amplify rather than mitigate
spurious focus on certain regions. Like most accuracy-oriented
post-processing techniques, AGAR could also be misused as a component
in larger pipelines for surveillance, automated content scraping, or
other applications whose downstream effects depend on context,
deployment, and policy choices outside the scope of this paper.
We therefore view AGAR as a research artifact: it should be evaluated
together with appropriate human oversight and use-case-specific safety
considerations before being deployed in any high-stakes setting.

\section{Prompt Used by AGAR}
\label{app:prompt}

AGAR uses a single short instruction template across all four backbones
(Qwen3-VL-8B, InternVL3.5-8B, GLM-4.1V-9B-Thinking, Glyph) and across
both the first (attention-reading) and second (re-rendered) forward
passes; the same template is also used by Plain VQA, so that AGAR's
gain over Plain VQA reflects the input image alone. The template asks
for a direct answer grounded in the visible text on the rendered page
and includes an explicit abstention option, and is shown in
Table~\ref{tab:agar_prompt}.

\begin{table}[h]
  \centering
  \small
  \caption{Prompt template used by AGAR (and Plain VQA).
  \texttt{\{question\}} is replaced by the dataset's natural-language
  question; nothing else in the prompt is altered between models or
  datasets.}
  \label{tab:agar_prompt}
  \begin{tabular}{p{0.92\linewidth}}
    \toprule
    \ttfamily Directly answer the question based on the text shown in
    the image, no explanation is needed. If the image does not contain
    any evidence, output ``I cannot answer based on the given context.''\\
    \ttfamily Question: \{question\} \\
    \bottomrule
  \end{tabular}
\end{table}


\clearpage
\section*{NeurIPS Paper Checklist}

\begin{enumerate}

\item {\bf Claims}
    \item[] Question: Do the main claims made in the abstract and introduction accurately reflect the paper's contributions and scope?
    \item[] Answer: \answerYes{} 
    \item[] Justification: The abstract accurately reflects the content in the intro, method, and experiments

\item {\bf Limitations}
    \item[] Question: Does the paper discuss the limitations of the work performed by the authors?
    \item[] Answer: \answerYes{} 
    \item[] Justification: See Conclusion and Limitation Section.

\item {\bf Theory assumptions and proofs}
    \item[] Question: For each theoretical result, does the paper provide the full set of assumptions and a complete (and correct) proof?
    \item[] Answer: \answerNA{} 
    \item[] Justification: No theory included

    \item {\bf Experimental result reproducibility}
    \item[] Question: Does the paper fully disclose all the information needed to reproduce the main experimental results of the paper to the extent that it affects the main claims and/or conclusions of the paper (regardless of whether the code and data are provided or not)?
    \item[] Answer: \answerYes{} 
    \item[] Justification: We fully disclose all the information needed to reproduce the main experimental results of the paper. We provide anonymous GitHub repos.

\item {\bf Open access to data and code}
    \item[] Question: Does the paper provide open access to the data and code, with sufficient instructions to faithfully reproduce the main experimental results, as described in supplemental material?
    \item[] Answer: \answerYes{} 
    \item[] Justification: We provide anonymous github repos

\item {\bf Experimental setting/details}
    \item[] Question: Does the paper specify all the training and test details (e.g., data splits, hyperparameters, how they were chosen, type of optimizer) necessary to understand the results?
    \item[] Answer: \answerYes{} 
    \item[] Justification: Please check Section Experiment for details

\item {\bf Experiment statistical significance}
    \item[] Question: Does the paper report error bars suitably and correctly defined or other appropriate information about the statistical significance of the experiments?
    \item[] Answer: \answerYes{} 
    \item[] Justification: Please check preliminaries and our experiments.

\item {\bf Experiments compute resources}
    \item[] Question: For each experiment, does the paper provide sufficient information on the computer resources (type of compute workers, memory, time of execution) needed to reproduce the experiments?
    \item[] Answer: \answerYes{} 
    \item[] Justification: Check  Experiment Section and Appendix.
    
\item {\bf Code of ethics}
    \item[] Question: Does the research conducted in the paper conform, in every respect, with the NeurIPS Code of Ethics \url{https://neurips.cc/public/EthicsGuidelines}?
    \item[] Answer: \answerYes{} 
    \item[] Justification: Yes.

\item {\bf Broader impacts}
    \item[] Question: Does the paper discuss both potential positive societal impacts and negative societal impacts of the work performed?
    \item[] Answer:  \answerYes{} 
    \item[] Justification: See Appendix and Conclusion.
    
\item {\bf Safeguards}
    \item[] Question: Does the paper describe safeguards that have been put in place for responsible release of data or models that have a high risk for misuse (e.g., pre-trained language models, image generators, or scraped datasets)?
    \item[] Answer: \answerNA{} 
    \item[] Justification: No risk.

\item {\bf Licenses for existing assets}
    \item[] Question: Are the creators or original owners of assets (e.g., code, data, models), used in the paper, properly credited and are the license and terms of use explicitly mentioned and properly respected?
    \item[] Answer: \answerYes{} 
    \item[] Justification: See Appendix.

\item {\bf New assets}
    \item[] Question: Are new assets introduced in the paper well documented and is the documentation provided alongside the assets?
    \item[] Answer: \answerYes{} 
    \item[] Justification: We do not release new datasets or pretrained models---all experiments use existing public benchmarks (NQ, HotpotQA, NewsQA, TriviaQA, LongBench, LoCoMo) under their original licenses. We do release the code for AGAR at an anonymous URL provided in the abstract; the repository includes a README with installation, usage, and reproduction instructions.

\item {\bf Crowdsourcing and research with human subjects}
    \item[] Question: For crowdsourcing experiments and research with human subjects, does the paper include the full text of instructions given to participants and screenshots, if applicable, as well as details about compensation (if any)? 
    \item[] Answer: \answerNA{} 
    \item[] Justification: Not applicable.

\item {\bf Institutional review board (IRB) approvals or equivalent for research with human subjects}
    \item[] Question: Does the paper describe potential risks incurred by study participants, whether such risks were disclosed to the subjects, and whether Institutional Review Board (IRB) approvals (or an equivalent approval/review based on the requirements of your country or institution) were obtained?
    \item[] Answer: \answerNA{}.
    \item[] Justification: Not applicable.

\item {\bf Declaration of LLM usage}
    \item[] Question: Does the paper describe the usage of LLMs if it is an important, original, or non-standard component of the core methods in this research? Note that if the LLM is used only for writing, editing, or formatting purposes and does \emph{not} impact the core methodology, scientific rigor, or originality of the research, declaration is not required.
    \item[] Answer: \answerYes{} 
    \item[] Justification: We use (V)LLM's internal signal to design our methods, see Section 4.

\end{enumerate}

\end{document}